\documentclass{article}
\PassOptionsToPackage{numbers, compress}{natbib}
\usepackage[final]{neurips_2021}

\usepackage[utf8]{inputenc} 
\usepackage[T1]{fontenc}    
\usepackage{graphicx}
\usepackage{amsmath,amssymb,amsthm,mathabx}
\usepackage{booktabs}
\usepackage{algorithmic}
\usepackage[linesnumbered,ruled,vlined]{algorithm2e}
\usepackage{acronym}
\usepackage{enumitem}
\usepackage[
    colorlinks, 
    linkcolor=green,
    anchorcolor=blue, 
    citecolor=blue
]{hyperref}
\usepackage{setspace}
\usepackage[skip=3pt,font=small]{subcaption}
\usepackage[skip=3pt,font=small]{caption}
\usepackage[dvipsnames]{xcolor}
\usepackage[capitalise]{cleveref}
\usepackage{tabularx,multirow,array,makecell}
\usepackage{overpic,wrapfig}

\usepackage{listings}
\lstdefinestyle{mystyle}{
  backgroundcolor=\color{backcolour},   commentstyle=\color{codegreen},
  keywordstyle=\color{magenta},
  numberstyle=\tiny\color{codegray},
  stringstyle=\color{codepurple},
  basicstyle=\ttfamily\footnotesize,
  commentstyle=\color{red!10!green!70}\textit,
  breakatwhitespace=false,         
  breaklines=true,                 
  captionpos=b,                    
  keepspaces=true,                 
  numbers=left,                    
  numbersep=5pt,                  
  showspaces=false,                
  showstringspaces=false,
  showtabs=false,                  
  tabsize=2
}

\makeatletter
\DeclareRobustCommand\onedot{\futurelet\@let@token\@onedot}
\def\@onedot{\ifx\@let@token.\else.\null\fi\xspace}
\def\eg{\emph{e.g}\onedot} 

\def\ie{\emph{i.e}\onedot}

\def\wrt{w.r.t\onedot}

\makeatother

\newcommand{\w}{\mathbf{w}}
\newcommand{\x}{\mathbf{x}}
\newcommand{\y}{\mathbf{y}}
\newcommand{\z}{\mathbf{z}}
\newcommand{\R}{\mathbb{R}}
\newcommand{\I}{\mathbf{I}}
\newcommand{\E}{\mathbf{E}}

\makeatletter
\renewcommand{\paragraph}{%
  \@startsection{paragraph}{4}%
  {\z@}{0ex \@plus 0ex \@minus 0ex}{-1em}%
  {\hskip\parindent\normalfont\normalsize\bfseries}%
}
\makeatother

\graphicspath{{figures/}}

\crefname{algocf}{alg.}{algs.}
\Crefname{algocf}{Algrithm}{Algrithm}

\definecolor{gblue}{HTML}{4285F4}
\definecolor{gred}{HTML}{DB4437}

\acrodef{drc}[DRC]{Deep Region Competition}
\acrodef{moe}[MoE]{Mixture of Experts}
\acrodef{mlem}[ML-EM]{Maximum-Likelihood Expectation-Maximization}
\acrodef{lebm}[LEBM]{Latent-space Energy-Based Model}
\acrodef{mle}[MLE]{Maximum Likelihood Estimation}
\acrodef{em}[EM]{Expectation-Maximization}
\acrodef{birds}[Birds]{Caltech-UCSD Birds-200-2011}
\acrodef{dogs}[Dogs]{Stanford Dogs}
\acrodef{cars}[Cars]{Stanford Cars}
\acrodef{tmds}[TM-dSprites]{Textured Multi-dSprites}
\acrodef{tvn}[TV-norm]{Total Variation norm}

\newcommand{\tabincell}[2]{\begin{tabular}{@{}#1@{}}#2\end{tabular}}

\title{Unsupervised Foreground Extraction via\\Deep Region Competition}

\author{
    Peiyu Yu$^1$ \\
    \texttt{yupeiyu98@g.ucla.edu}
    \And Sirui Xie$^1$ \\
    \texttt{srxie@ucla.edu}
    \And Xiaojian Ma$^1$ \\
    \texttt{xiaojian.ma@ucla.edu}
    \And Yixin Zhu$^{3}$ \\
    \texttt{y@bigai.ai}
    \And Ying Nian Wu$^2$ \\
    \texttt{ywu@stat.ucla.edu}
    \And Song-Chun Zhu$^{1, 2, 3}$ \\
    \texttt{sczhu@stat.ucla.edu}
    \AND
    \normalfont{$^1$UCLA Department of Computer Science\quad{}}
    \normalfont{$^2$UCLA Department of Statistics}\\
    $^3$Beijing Institute for General Artificial Intelligence (BIGAI)
}

\begin{document}

\maketitle

\begin{abstract}
We present \ac{drc}, an algorithm designed to extract foreground objects from images in a fully unsupervised manner.
Foreground extraction can be viewed as a special case of generic image segmentation that focuses on identifying and disentangling objects from the background.
In this work, we rethink the foreground extraction by reconciling energy-based prior with generative image modeling in the form of \ac{moe}, where we further introduce the learned pixel re-assignment as the essential inductive bias to capture the regularities of background regions. With this modeling, the foreground-background partition can be naturally found through \ac{em}.
We show that the proposed method effectively exploits the interaction between the mixture components during the partitioning process, which closely connects to region competition~\cite{zhu1996region}, a seminal approach for generic image segmentation. Experiments demonstrate that \ac{drc} exhibits more competitive performances on complex real-world data and challenging multi-object scenes compared with prior methods. Moreover, we show empirically that \ac{drc} can potentially generalize to novel foreground objects even from categories unseen during training.\footnote{Code and data available at \href{https://github.com/yuPeiyu98/Deep-Region-Competition}{https://github.com/yuPeiyu98/Deep-Region-Competition}}
\end{abstract}

\section{Introduction}

Foreground extraction, being a special case of generic image segmentation, aims for a binary partition of the given image with specific semantic meaning, \ie, a foreground that typically contains identifiable objects and the possibly less structural remaining regions as the background.
There is a rich literature on explicitly modeling and representing a given image as foreground and background (or more general visual regions), such that a generic inference algorithm can produce plausible segmentations ideally for any images without or with little supervision~\cite{zhu1996region,shi2000normalized,tu2002image,boykov2001interactive,rother2004grabcut,cheng2014global,jiang2013salient,zhu2014saliency}.
However, such methods essentially rely on low-level visual features (\eg, edges, color, and texture), and some further require human intervention at initialization~\cite{boykov2001interactive,rother2004grabcut}, which largely limits their practical performance on modern datasets of complex natural images with rich semantic meanings~\cite{lin2014microsoft,everingham2010pascal}. These datasets typically come with fine-grained semantic annotations, exploited by supervised methods that learn representation and inference algorithm as one monolithic network~\cite{zhao2017pyramid,long2015fully,badrinarayanan2017segnet,chen2017deeplab,ronneberger2015u,he2017mask}. Despite the success of densely supervised learning, the unsupervised counterpart is still favored due to its resemblance to how humans perceive the world~\cite{chater2006probabilistic,shipley2001fragments}.

Attempting to combine unsupervised or weakly supervised learning with modern neural networks, three lines of work surge recently for foreground extraction: (1) deep networks as feature extractors for canonical segmentation algorithms, (2) GAN-based foreground-background disentanglement, and (3) compositional latent variable models with slot-based object modeling. Despite great successes of these methods, the challenge of unsupervised foreground extraction remains largely open.

Specifically, the first line of work trains designated deep feature extractors for canonical segmentation algorithms or metric networks as learned partitioning criteria~\cite{xia2017w,kanezaki2018unsupervised,ji2019invariant}. These methods (\eg, W-Net~\cite{xia2017w}) define foreground objects' properties using learned features or criteria and are thus generally bottle-necked by the selected post-processing segmentation algorithm~\cite{arbelaez2010contour,achanta2012slic}. 
As a branch of pioneering work that moves beyond these limitations,~\citet{yang2019unsupervised, yang2021dystab} have recently proposed a general contextual information separation principle and an efficient adversarial learning method that is generally applicable to unsupervised segmentation, separation and detection. 
GAN-based models~\cite{goodfellow2014generative,yang2017lr,chen2019unsupervised,ostyakov2018seigan,singh2019finegan,benny2020onegan} capture the foreground objectness with oversimplified assumptions or require additional supervision to achieve foreground-background disentanglement. 
For example, the segmentation model in ReDO~\cite{chen2019unsupervised} is trained by redrawing detected objects, which potentially limits its application to datasets with diverse object shapes. OneGAN~\cite{benny2020onegan} and its predecessors~\cite{ostyakov2018seigan,singh2019finegan}, though producing impressive results on foreground extraction, require a set of background images without foreground objects as additional inputs.
Lastly, compositional latent variable models~\cite{greff2016tagger,eslami2016attend,greff2017neural,van2018relational,burgess2019monet,greff2019multi,locatello2020object,engelcke2020genesis,lin2020space} include the background as a ``virtual object'' and induce the independence of object representations using an identical generator for all object slots. Although these methods exhibit strong performance on synthetic multi-object datasets with simple backgrounds and foreground shapes, they may fail on complex real-world data or even synthetic datasets with more challenging backgrounds~\cite{greff2019multi,locatello2020object}. In addition, few unsupervised learning methods have provided explicit identification of foreground objects and background regions. While they can generate valid segmentation masks, most of these methods do not specify which output corresponds to the foreground objects.
These deficiencies necessitate rethinking the problem of unsupervised foreground extraction. We propose to confront the challenges in formulating (1) a generic inductive bias for modeling foreground and background regions that can be baked into neural generators, and (2) an effective inference algorithm based on a principled criterion for foreground-background partition.

Inspired by Region Competition~\cite{zhu1996region}, a seminal approach that combines optimization-based inference~\cite{kass1988snakes,cohen1991active,adams1994seeded} and probabilistic visual modeling~\cite{zhu1998filters,guo2007primal} by minimizing a generalized Bayes criterion~\cite{leclerc1989constructing}, we propose to solve the foreground extraction problem by reconciling energy-based prior~\cite{pang2020learning} with generative image modeling in the form of \acf{moe}~\cite{jacobs1991adaptive,jordan1994hierarchical}. To generically describe background regions, we further introduce the learned pixel re-assignment as the essential inductive bias to capture their regularities. Fueled by our modeling, we propose to find the foreground-background partition through \acf{em}. Our algorithm effectively exploits the interaction between the mixture components during the partitioning process, echoing the intuition described in Region Competition~\cite{zhu1996region}. We therefore coin our method \acf{drc}. We summarize our \textbf{contributions} as follows: 
\begin{enumerate}[leftmargin=*]
    \item We provide probabilistic foreground-background modeling by reconciling energy-based prior with generative image modeling in the form of \ac{moe}. With this modeling, the foreground-background partition can be naturally produced through \ac{em}. We further introduce an inductive bias, \emph{pixel re-assignment}, to facilitate foreground-background disentanglement. 
    \item In experiments, we demonstrate that \ac{drc} exhibits more competitive performances on complex real-world data and challenging multi-object scenes compared with prior methods. Furthermore, we empirically show that using learned pixel re-assignment as the inductive bias helps to provide explicit identification for foreground and background regions.
    \item We find that \ac{drc} can potentially generalize to novel foreground objects even from categories unseen during training, which may provide some inspiration for the study of out-of-distribution (OOD) generalization in more general unsupervised disentanglement.
\end{enumerate}

\section{Related Work}

A typical line of methods frames unsupervised or weakly supervised foreground segmentation within a generative modeling context. Several methods build upon generative adversarial networks (GAN)~\cite{goodfellow2014generative} to perform foreground segmentation. LR-GAN~\cite{yang2017lr} learns to generate background regions and foreground objects separately and recursively, which simultaneously produces the foreground objects mask. ReDO (ReDrawing of Objects)~\cite{chen2019unsupervised} proposes a GAN-based object segmentation model, based on the assumption that replacing the foreground object in the image with a generated one does not alter the distribution of the training data, given that the foreground object is correctly discovered. Similarly, SEIGAN~\cite{ostyakov2018seigan} learns to extract foreground objects by recombining the foreground objects with the generated background regions. FineGAN~\cite{singh2019finegan} hierarchically generates images (\ie, first specifying the object shape and then the object texture) to disentangle the background and foreground object. \citet{benny2020onegan} further hypothesize that a method solving an ensemble of unsupervised tasks altogether improves the model performance compared with the one that solves each individually. Therefore, they train a complex GAN-based model (OneGAN) to solve several tasks simultaneously, including foreground segmentation. Although LR-GAN and FineGAN do produce masks as part of their generative process, they cannot segment a given image. Despite SEIGAN and OneGAN achieving decent performance on foreground-background segmentation, these methods require a set of clean background images as additional inputs for weak supervision. ReDO captures the foreground objectness with possibly oversimplified assumptions, limiting its application to datasets with diverse object shapes.

On another front, compositional generative scene models~\cite{greff2016tagger,eslami2016attend,greff2017neural,van2018relational,burgess2019monet,greff2019multi,locatello2020object,engelcke2020genesis,lin2020space}, sharing the idea of scene decomposition stemming from DRAW~\cite{gregor2015draw}, learn to represent foreground objects and background regions in terms of a collection of latent variables with the same representational format. These methods typically exploit the spatial mixture model for generative modeling. Specifically, IODINE~\cite{greff2019multi} proposes a slot-based object representation method and models the latent space using iterative amortized inference~\cite{marino2018iterative}. Slot-Attention~\cite{locatello2020object}, as a step forward, effectively incorporates the attention mechanism into the slot-based object representation for flexible foreground object binding. Both methods use fully shared parameters among individual mixture components to entail permutation invariance of the learned multi-object representation. Alternative models such as MONet~\cite{burgess2019monet} and GENESIS~\cite{engelcke2020genesis} use multiple encode-decode steps for scene decomposition and foreground object extraction. Although these methods exhibit strong performance on synthetic multi-object datasets with simple background and foreground shapes, they may fail when dealing with complex real-world data or even synthetic datasets with more challenging background~\cite{greff2019multi,locatello2020object}.

More closely related to the classical methods, another line of work focuses on utilizing image features extracted by deep neural networks or designing energy functions based on data-driven methods to define the desired property of foreground objects. \citet{pham2018scenecut} and \citet{silberman2012indoor} obtain impressive results when depth images are accessible in addition to conventional RGB images, while such methods are not directly applicable for data with RGB images alone. W-Net~\cite{xia2017w} extracts image features via a deep auto-encoder jointly trained by minimizing reconstruction error and normalized cut. The learned features are further processed by CRF smoothing to perform hierarchical segmentation. \citet{kanezaki2018unsupervised} proposes to employ a neural network as part of the partitioning criterion (inspired by \citet{ulyanov2020deep}) to minimize the chosen intra-region pixel distance for segmentation directly. \citet{ji2019invariant} propose to use Invariant Information Clustering as the objective for segmentation, where the network is trained to be part of the learned distance. As an interesting extension, one may also consider adapting methods that automatically discover object structures~\cite{lorenz2019unsupervised} to foreground extraction. Though being pioneering work in image segmentation, the aforementioned methods are generally bottle-necked by the selected post-processing segmentation algorithm or require extra transformations to produce meaningful foreground segmentation masks. 
\citet{yang2019unsupervised, yang2021dystab} in their seminal work propose an information-theoretical principle and adversarial contextual model for unsupervised segmentation and detection by partitioning images into maximally independent sets, with the objective of minimizing the predictability of one set by the other sets. Additional efforts have also been devoted to weakly supervised foreground segmentation using image classification labels~\cite{papandreou2015weakly,pathak2015constrained,huang2018weakly}, bounding boxes~\cite{dai2015boxsup,khoreva2017simple}, or saliency maps~\cite{oh2017exploiting,zeng2019joint, voynov2021object}.

\section{Methodology}
Foreground extraction performs a binary partition for the image $\I$ to extract the foreground region. Without explicit supervision, we propose to use learned pixel re-assignment as a generic inductive bias for background modeling, upon which we derive an \ac{em}-like partitioning algorithm. Compared with prior methods, our algorithm can handle images with more complex foreground shapes and background patterns, while providing explicit identification of foreground and background regions.

\setstretch{1}

\subsection{Preliminaries}\label{sec:lebm}

Adopting the language of \ac{em} algorithm, we assume that for the observed sample $\x \in \R^D$, there exists $\z \in \R^d$ as its latent variables. The complete-data distribution is
\begin{equation}
    p_\theta(\z,\x) = p_\alpha(\z)p_\beta(\x|\z),   
\end{equation}
where $p_\alpha(\z)$ is the prior model with parameters $\alpha$, $p_\beta(\x|\z)$ is the top-down generative model with parameters $\beta$, and $\theta = (\alpha,\beta)$.

The prior model $p_\alpha(\z)$ can be formulated as an energy-based model, which we refer to as the \ac{lebm}~\cite{pang2020learning} throughout the paper:
\begin{equation}
    p_\alpha(\z) = \frac{1}{Z_{\alpha}}\exp
                    \left(
                        f_\alpha(\z)
                    \right)p_0(\z),
    \label{equ:lebm_def}
\end{equation}
where $f_\alpha(\z)$ can be parameterized by a neural network, $Z_\alpha$ is the partition function, and $p_0(\z)$ is a reference distribution, assumed to be isotropic Gaussian prior commonly used for the generative model. The prior model in \cref{equ:lebm_def} can be interpreted as an energy-based correction or exponential tilting of the original prior distribution $p_0$.

The \ac{lebm} can be learned by \ac{mle}. Given a training sample $\x$, the learning gradient for $\alpha$ is derived as shown by \citet{pang2020learning},
\begin{equation}
    \delta_\alpha (\x) = \E_{p_\theta(\z|\x)} \left[\nabla_\alpha f_\alpha(\z) \right]
                       - \E_{p_\alpha(\z)} \left[\nabla_\alpha f_\alpha(\z) \right].
    \label{equ:lebm_learn}
\end{equation}

In practice, the above expectations can be approximated by Monte-Carlo average, which requires sampling from $p_\theta(\z|\x)$ and $p_\alpha(\z)$. This step can be done with stochastic gradient-based methods, such as Langevin dynamics~\cite{welling2011bayesian} or Hamiltonian Monte Carlo~\cite{brooks2011handbook}.

An extension to \ac{lebm} is to further couple the vector representation $\z$ with a symbolic representation $\y$~\cite{pang2021latent}. Formally, $\y$ is a K-dimensional one-hot vector, where $K$ is the number of possible $\z$ categories. Such symbol-vector duality can provide extra entries for auxiliary supervision; we will detail it in \cref{sec:techd}.

\subsection{Generative Image Modeling}\label{sec:model}

\paragraph{\acf{moe} for Image Generation}

\begin{figure*}[t!]
    \centering
    \includegraphics[width=\linewidth]{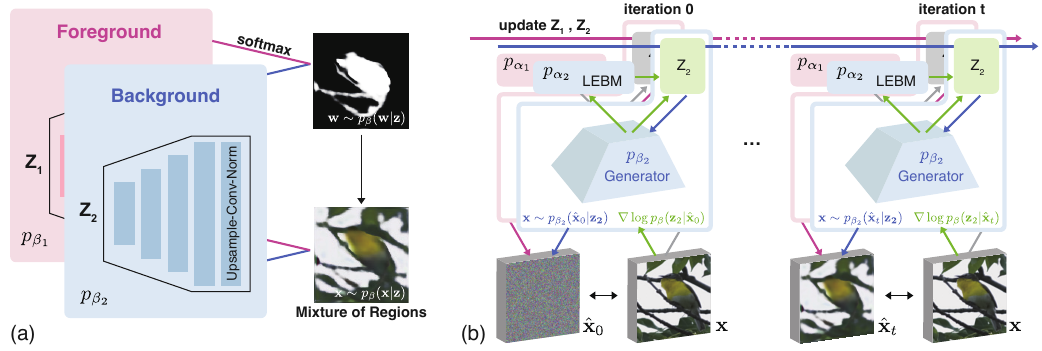}
    \caption{\textbf{Overview of \ac{drc}.} (a) The model generates foreground and background regions using sampled latent variables $\z = \{\z_1, \z_2\}$. $p_{\beta_k},~k=1,2$ represents the generator for each region. Of note, the pixel re-assignment function is absorbed in the background generator; see \cref{sec:model} for details. (b) \ac{drc} samples the latent variables $\z$ in an iterative manner. Let $\x$ denote the observed image; we use $\hat{\x}_t,~t=0,1,...$ to represent the image generated by $p_\beta(\x | \z)$ at the $t$-th sampling step. \ac{drc} has a two-step workflow for learning unsupervised foreground extractors that resembles the E- and M-step in the classic \ac{em} algorithm. In the E-step, it employs gradient-based MCMC sampling to infer the latent variables $\z$ as shown in (b). Of note, only the latent variables $\z$ are updated in this step. In the M-step, the sampled latent variables $\z$ are fed into the model for image generation as shown in (a), where the generators are updated to minimize the reconstruction error.}
    \label{fig:overview}
\end{figure*}

Inspired by the regional homogenity assumption proposed by \citet{zhu1996region}, we use separate priors and generative models for foreground and background regions, indexed as $\alpha_k$ and $\beta_k,~k=1,2$, respectively; see \cref{fig:overview}. This design leads to the form of \ac{moe}~\cite{jacobs1991adaptive,jordan1994hierarchical} for image modeling, as shown below.

Let us start by considering only the i-th pixel of the observed image $\x$, denoted as $\x_i$. We use a binary one-hot random variable $\w_i$ to indicate whether the i-th pixel belongs to the foreground region. Formally, we have $\w_i = [w_{i1}, w_{i2}]$, $w_{ik} \in \{0, 1\}$ and $\sum_{k=1}^{2}w_{ik} = 1$. Let $w_{i1}=1$ indicate that the i-th pixel $\x_i$ belongs to the foreground, and $w_{i2}=1$ indicate the opposite.

We assume that the distribution of $\w_i$ is prior-dependent. Specifically, the mixture parameter $\pi_{ik},~k=1,2$, is defined as the output of a gating function $ \pi_{ik} = p_{\beta}(w_{ik}=1 | \z) = \text{Softmax}(l_{ik}) $; $l_{ik} = h_{\beta_k}(\z_k),~k=1,2$ are the logit scores given by the foreground and background generative models respectively; $\beta = \{\beta_1, \beta_2\}, \z = \{\z_1, \z_2\}$. Taken together, the joint distribution of $\w_i$ is
\begin{equation}
    p_\beta(\w_{i} | \z) = \prod\limits_{k=1}^2 \pi_{ik}^{w_{ik}}.
\end{equation}

The learned distribution of foreground and background contents are 
\begin{equation}
    p_{\beta}(\x_i | w_{ik}=1, \z_k) = p_{\beta_k}(\x_i | \z_k) 
                                     \sim \mathbf{N}(g_{\beta_k}(\z_k), \sigma^2 \mathbf{I}),
                                     ~k = 1, 2
\end{equation}
where we assume that the generative model for region content, $p_{\beta_k}(\x_i | \z_k),~k=1,2$, follows a Gaussian distribution parameterized by the generator network $g_{\beta_k}$. As in VAE, $\sigma$ takes an assumed value. We follow the common practice and use a shared generator for parameterizing $\pi_{ik}$ and $p_{\beta_k}(\x_i | \z_k)$. We use separate branches only at the output layer to generate logits and contents.

Generating $\x_i$ based on $\w_i$'s distribution involves two steps: (1) sample $\w_i$ from the distribution $p_\beta(\w_{i} | \z)$, and (2) choose either the foreground model (\ie, $p_{\beta_1}(\x_i | \z_1)$) or the background model (\ie, $p_{\beta_2}(\x_i | \z_2)$) to generate $\x_i$ based on the sampled $\w_i$. As such, this distribution of $\x_i$ is a \ac{moe},
\begin{equation}
    p_\beta(\x_i | \z) 
                = \sum\limits_{k=1}^{2}p_\beta(w_{ik}=1 | \z) 
                                       p_{\beta}(\x_i | w_{ik}=1, \z_k)
                = \sum\limits_{k=1}^2 \pi_{ik}p_{\beta_k}(\x_i | \z_k),
\end{equation}
wherein the posterior responsibility of $w_{ik}$ is 
\begin{equation}
    \gamma_{ik} = p(w_{ik} = 1 | \x_i, \z)
                = \frac{\pi_{ik}p_{\beta_k}( \x_i | \z_k)}
                       {\sum_{m=1}^2\pi_{im}p_{\beta_m}(\x_i | \z_m)}, ~k=1,2.
    \label{equ:post_w}
\end{equation}
Using a fully-factorized joint distribution of $\x$, we have
$   
    p_\beta(\x | \z) = \prod_{i=1}^{D}\sum_{k=1}^2 \pi_{ik}p_{\beta_k}(\x_i | \z_k) 
$
as the generative modeling of $\x \in \R^D$.

\setlength\floatsep{0.2\baselineskip plus 3pt minus 2pt}
\setlength\textfloatsep{0.2\baselineskip plus 3pt minus 2pt}
\setlength\dbltextfloatsep{0.2\baselineskip plus 3pt minus 2 pt}
\setlength\intextsep{0.2\baselineskip plus 3pt minus 2 pt}

\begin{wrapfigure}{r}{0.42\linewidth}
    \vspace{-15pt}
    \centering
    \includegraphics[width=\linewidth]{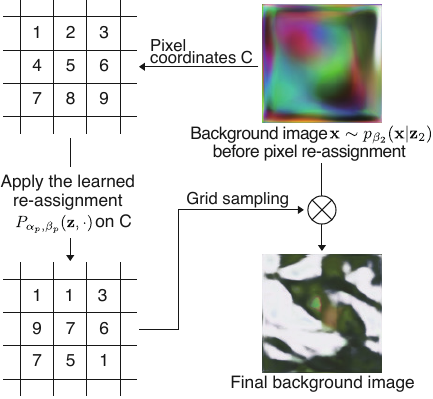}
    \caption{\textbf{Pixel re-assignment.} The output of $\beta_p$ can be viewed as a learned re-assignment of the original background pixels that follows the mapped grid $P_{\alpha_p, \beta_p}(\z, C)$. Note that the re-assignment function $P_{\alpha_p, \beta_p}(\z, \cdot)$ might not be injective. The final background image is generated via grid sampling. }
    \label{fig:pix_perm}
\end{wrapfigure}

\paragraph{Learning Pixel Re-assignment for Background Modeling}

We use pixel re-assignment in the background generative model as the essential inductive bias for modeling the background region. This is partially inspired by the concepts of ``texture'' and ``texton'' by Julez~\cite{guo2007primal,julesz1981textons}, where the textural part of an image may contain fewer structural elements in preattentive vision, which coincides with our intuitive observation of the background regions.

We use a separate pair of energy-based prior model $\alpha_{\texttt{pix}}$ and generative model $\beta_{\texttt{pix}}$ to learn the re-assignment. For simplicity, we absorb $\alpha_{\texttt{pix}}$ and $\beta_{\texttt{pix}}$ in the models for background modeling, \ie, $\alpha_2$ and $\beta_2$, respectively. In practice, the re-assignment follows the output of $\beta_{\texttt{pix}}$, a shuffling grid with the same size of the image $\x$. Its values indicate the re-assigned pixel coordinates; see \cref{fig:pix_perm}. We find that shuffling the background pixels using the learned re-assignment facilitates the model to capture the regularities of the background regions. Specifically, the proposed model with this essential inductive bias learns to constantly give the correct mask assignment, whereas most previous fully unsupervised methods do not provide explicit identification of the foreground and background regions; see discussion in \cref{sec:results} for more details.

\setstretch{1}

\subsection{Deep Region Competition: from Generative Modeling to Foreground Extraction}\label{sec:drc}

The complete-data distribution from the image modeling is
\begin{equation}
    \begin{aligned}
        p_{\theta}(\x, \z, \w) 
            &= p_{\beta}(\x | \w, \z)p_{\beta}(\w | \z)p_\alpha(\z) \\
            &=  \left(
                    \prod\limits_{i=1}^{D} \prod\limits_{k=1}^{2} p_{\beta_k}(\x_i | \z_k)^{w_{ik}}
                \right)
                \left(
                    \prod\limits_{i=1}^{D} \prod\limits_{k=1}^{2} \pi_{ik}^{w_{ik}}
                \right)
                p_\alpha(\z) \\
            &=  p_\alpha(\z)
                \prod\limits_{i=1}^{D} \prod\limits_{k=1}^{2} 
               \left( 
                \pi_{ik} p_{\beta_k}(\x_i | \z_k) 
               \right)^{w_{ik}},
    \end{aligned}
\end{equation}
where $p_\alpha(\z) = p_{\alpha_1}(\z_1)p_{\alpha_2}(\z_2)$ is the prior model given by \acp{lebm}. $\alpha = \{\alpha_1, \alpha_2\}$, and $\theta = \{\alpha, \beta\}$. $\w$ is the vector of $(\w_i), i=1, ... D$, whose joint distribution is assumed to be fully-factorized. 

Next, we derive the complete-data log-likelihood as our learning objective:
\begin{equation}
    \mathcal{L}(\theta) = 
    \log p_{\theta}(\x, \z, \w) = \log p_\alpha(\z) + 
        \sum\limits_{i=1}^D\sum\limits_{k=1}^2
            w_{ik}\left(
                \log \pi_{ik} + \log p_{\beta_k}(\x_i | \z_k)
            \right).
    \label{equ:complete_lkhd}
\end{equation}
Of note, $\w$ and $\z$ are unobserved variables in the modeling, which makes it impossible to learn the model directly through \ac{mle}. To calculate the gradients of $\theta$, we instead optimize $\mathbf{E}_{\z \sim p(\z | \x), \w \sim p(\w | \x, \z)} [\mathcal{L}(\theta)]$ based on the fact that underlies the \ac{em} algorithm:
\begin{equation}
    \begin{aligned}
        \nabla_{\theta}\log p_\theta(\x) 
            &= \int_{\z} p_\theta(\z | \x)d\z \int_{\w} p_\theta(\w | \z, \x) \nabla_\theta \log p_{\theta}(\x, \z, \w) d\w \\
            &= \mathbf{E}_{\z \sim p_\theta(\z | \x), \w \sim p_\theta(\w | \x, \z)}[\nabla_\theta \log p_{\theta}(\x, \z, \w) ].
    \end{aligned}
\end{equation} 

Therefore, the derived surrogate learning objective becomes
\begin{equation}
        \max\limits_{\theta}~ \mathbf{E}_{\z \sim p_\theta(\z|\x)} 
            \left[
                \mathcal{J}(\theta) 
            \right], ~
        \text{s.t.} ~\forall i, \sum\limits_{k=1}^2 \pi_{ik} = 1,
    \label{equ:surr_obj}
\end{equation}
\begin{equation}
    \mathcal{J}(\theta) = 
           \enskip \underbrace{
                \log p_\alpha(\z)
            }_{\text{objective for \ac{lebm}}}
          +\enskip \underbrace{
                \sum\limits_{i=1}^D\sum\limits_{k=1}^2 \gamma_{ik}\log \pi_{ik}
            }_{\text{foreground-background partitioning}}
          +\enskip \underbrace{
                \sum\limits_{i=1}^D\sum\limits_{k=1}^2 \gamma_{ik}\log p_{\theta_k}(\x_i | \z_k)
            }_{\text{objective for image generation}},
    \label{equ:cond_exp}
\end{equation}
where $\mathcal{J}(\theta)=\mathbf{E}_{\w \sim p_\theta(\w | \x, \z)}\left[\mathcal{L}(\theta)\right]$ is the conditional expectation of $\w$, which can be calculated in closed form; see the supplementary material for additional details.

\cref{equ:surr_obj} has an intuitive interpretation. We can decompose the learning objective into three components as in \cref{equ:cond_exp}. In particular, the second term $\sum_{i=1}^D\sum_{k=1}^2 \gamma_{ik}\log \pi_{ik}$ has a similar form to the cross-entropy loss commonly used for supervised segmentation task, where the posterior responsibility $\gamma_{ik}$ serves as the target distribution. It is as if the foreground and background generative models compete with each other to fit the distribution of each pixel $\x_i$. If the pixel value at $\x_i$ fits better to the distribution of foreground, $p_{\beta_1}(\x_i | \z_1)$, than to that of background, $p_{\beta_2}(\x_i | \z_2)$, the model tends to assign that pixel to the foreground region (see \cref{equ:post_w}), and vice versa. This mechanism is similar to the process derived in \citet{zhu1996region}, which is the reason why we coin our method \acf{drc}.

Prior to our proposal, several methods~\cite{zhu1996region,greff2019multi,locatello2020object} also employ mixture models and competition among the components to perform unsupervised foreground or image segmentation. The original Region Competition~\cite{zhu1996region} combines several families of image modeling with Bayesian inference but is limited by the expressiveness of the pre-specified probability distributions. More recent methods, including IODINE~\cite{greff2019multi} and Slot-attention~\cite{locatello2020object}, learn amortized inference networks for latent variables and induce the independence of foreground and background representations using an identical generator. Our method combines the best of the two worlds, reconciling the expressiveness of learned generators with the regularity of generic texture modeling under the framework of \ac{lebm}.

To optimize the learning objective in \cref{equ:surr_obj}, we approximate the expectation by sampling from the prior $p_\alpha(\z)$ and posterior model $p_\theta(\z|\x) \propto p_\alpha(\z)p_\beta(\x|\z)$, followed by calculating the Monte Carlo average. We use Langevin dynamics~\cite{welling2011bayesian} to draw persistent MCMC samples, which iterates
\begin{equation}
    \z_{t+1} = \z_{t} + s\nabla_\z \log Q(\z_t) + \sqrt{2s} \epsilon_t,
    \label{equ:langevin_dyna}
\end{equation}
where $t$ is the Langevin dynamics's time step, $s$ the step size, and $\epsilon_t$ the Gaussian noise. $Q(\z)$ is the target distribution, being either $p_\alpha(\z)$ or $p_\theta(\z | \x)$. $\nabla_\z \log Q(\z_t)$ is efficiently computed via automatic differentiation in modern learning libraries~\cite{paszke2019pytorch}. We summarize the above process in \cref{alg:deep_region_comp}.

\begin{algorithm}[!htbp]
    \small
    \caption{\textbf{Learning models of \ac{drc} via \ac{em}.}}
    \label{alg:deep_region_comp}
	\KwIn{Learning iterations $T$, initial parameters for LEBMs $\alpha^{(0)} =\{ \alpha_1^{(0)}, \alpha_2^{(0)} \}$ and generators $\beta^{(0)} =\{ \beta_1^{(0)}, \beta_2^{(0)} \}$,
	$\theta^{(0)} = \{\alpha^{(0)}, \beta^{(0)}\}$, learning rate $\eta_\alpha$ for
	LEBMs, $\eta_\beta$ for foreground and background generators,
	observed examples $\{\x^{(i)}\}_{i=1}^N$, batch size $M$, and initial latent variables $\{\z_{-}^{(i)} = \{\z_{1-}^{(i)}, \z_{2-}^{(i)} \} \sim p_0(\z)\}_{i=1}^N$ and $\{\z_{+}^{(i)} = \{\z_{1+}^{(i)}, \z_{2+}^{(i)} \} \sim p_0(\z)\}_{i=1}^N$.}
    \KwOut{$\theta^{(T)} = \{\alpha_1^{(T)}, \beta_1^{(T)}, \alpha_2^{(T)}, \beta_2^{(T)}\}$.}
	\BlankLine
	
	\For{$t=0:T-1$}{
		Sample a minibatch of data $\{\x^{(i)}\}_{i=1}^M$;
		
	    \textbf{Prior sampling for learning LEBMs:} For each $\x^{(i)}$, update $\z_{-}^{(i)}$ using \cref{equ:langevin_dyna}, with target distribution $\pi(\z) = p_{\alpha^{(t)}}(\z)$;
	    
	    \textbf{Posterior sampling for foreground and background generation:} For each $\x^{(i)}$, update $\z_{+}^{(i)}$ using \cref{equ:langevin_dyna}, with target distribution $Q(\z) = p_{\theta^{(t)}}(\z|\x)$;
	    
	    \textbf{Update LEBMs: } 
	    $\alpha^{(t+1)} = \alpha^{(t)} + \eta_\alpha \frac{1}{m} 
	                    \sum_{i=1}^m [
	                        \nabla_\alpha f_{\alpha^{(t)}}(\z_{+}^{(i)}) -
	                        \nabla_\alpha f_{\alpha^{(t)}}(\z_{-}^{(i)})
	                    ]$;
		
		\textbf{Update foreground and background generators:}
		$\beta^{(t+1)} = \beta^{(t)} + \eta_\beta \frac{1}{m} 
	                    \sum_{i=1}^m 
	                        \nabla_\beta \log p_{\beta^{(t)}} (\x^{(i)} | \z_{+}^{(i)}) 
	                    $;
	}
\end{algorithm}

During inference, we initialize the latent variables $\z$ for MCMC sampling from Gaussian white noise and run only the posterior sampling step to obtain $\z_{+}$. The inferred mask and region images are then given by the outputs of generative models $p_{\beta_k}(\w | \z_{+})$ and $p_{\beta_k}(\x | \z_{+}),~k=1,2$, respectively.

\subsection{Technical Details}\label{sec:techd}

\paragraph{Pseudo label for additional regularization}

Although the proposed \ac{drc} explicitly models the interaction between the regions, it is still possible that the model converges to a trivial extractor, which treats the entire image as the foreground or background region, leaving the other region null. We exploit the symbolic vector $\y$ emitted by the \ac{lebm} (see \cref{sec:lebm}) for additional regularization. The strategy is similar to the mutual information maximization used in InfoGAN~\cite{chen2016infogan}. Specifically, we use the symbolic vector $\y$ inferred from $\z$ as the pseudo-class label for $\z$ and train an auxiliary classifier jointly with the above models; it ensures that the generated regions $\x_k$ contain similar symbolic information for $\z_k$. Intuitively, this loss prevents the regions from converging to null since the symbolic representation $\y_k$ would never be well retrieved if that did happen.

\paragraph{Implementation}

We adopt a similar architecture for the generator as in DCGAN~\cite{radford2015unsupervised} throughout the experiments and only change the dimension of the latent variables $\z$ for different datasets. The generator consists of a fully connected layer followed by five stacked upsample-conv-norm layers. We replace the batch-norm layers~\cite{ioffe2015batch} with instance-norm~\cite{ulyanov2016instance} in the architecture. The energy-term in \ac{lebm} is parameterized by a 3-layered MLP. We adopt orthogonal initialization~\cite{saxe2013exact} commonly used in generative models to initialize the networks and orthogonal regularization~\cite{brock2016neural} to facilitate training. In addition, we observe performance improvement when adding Total-Variation norm~\cite{rudin1992nonlinear} for the background generative model. More details, along with specifics of the implementation used in our experiments, are provided in the supplementary material.

\section{Experiments}\label{sec:exp}

We design experiments to answer three questions: (1) How does the proposed method compare to previous state-of-the-art competitors? (2) How do the proposed components contribute to the model performance? (3) Does the proposed method exhibit generalization on images containing unseen instances (\ie, same category but not the same instance) and even objects from novel categories?

To answer these questions, we evaluate our method on five challenging datasets in two groups: (1) \ac{birds}~\cite{cub}, \ac{dogs}~\cite{khosla2011novel}, and \ac{cars}~\cite{krause20133d} datasets; (2) CLEVR6~\cite{johnson2017clevr} and \ac{tmds}~\cite{dsprites17} datasets. The first group of datasets covers complex real-world domains, whereas the second group features environments of the multi-object foreground with challenging spatial configurations or confounding backgrounds. As to be shown, the proposed method is generic to various kinds of input and produces more competitive foreground-background partition results than prior methods.

\subsection{Results on Foreground Extraction}\label{sec:results}

\begin{table}[!htbp]
    \centering
    \begin{tabular}{lcccccccccc}
        \toprule
        \multicolumn{1}{c}{ } & 
        \multicolumn{6}{c}{Single Object} & 
        \multicolumn{4}{c}{Multi-Object} \\
        \cmidrule(r){2-7} \cmidrule(r){8-11} 
        
        \multicolumn{1}{c}{Model} & 
        \multicolumn{2}{c}{\ac{birds}} &
        \multicolumn{2}{c}{\ac{dogs}} &
        \multicolumn{2}{c}{\ac{cars}} &
        \multicolumn{2}{c}{CLEVR6} &
        \multicolumn{2}{c}{\ac{tmds}} \\
        \cmidrule(r){2-3} \cmidrule(r){4-5} \cmidrule(r){6-7} 
        \cmidrule(r){8-9} \cmidrule(r){10-11} 
        
        \multicolumn{1}{c}{ } & 
        \multicolumn{1}{c}{IoU} & 
        \multicolumn{1}{c}{Dice} &
        \multicolumn{1}{c}{IoU} & 
        \multicolumn{1}{c}{Dice} &
        \multicolumn{1}{c}{IoU} & 
        \multicolumn{1}{c}{Dice} &
        \multicolumn{1}{c}{IoU} & 
        \multicolumn{1}{c}{Dice} &
        \multicolumn{1}{c}{IoU} & 
        \multicolumn{1}{c}{Dice}\\
        \hline
        $\text{W-Net}^{*}$ & 24.8 & 38.9 & 47.7 & 62.1 & 52.8 & 67.6 & - & - & - & -\\
        GrabCut & 30.2 & 42.7 & 58.3 & 70.9 & 61.3 & 73.1 & 19.0 & 30.5 & 61.9 & 71.0 \\
        \hline
        $\text{ReDO}^\S$ & 46.5 & 60.2 & 55.7 & 70.3 & 52.5 & 68.6 & 18.6 & 31.0 & 9.4 & 17.2 \\
        $\text{OneGAN}^{*\dag}$ & 55.5 & 69.2 & 71.0 & 81.7 & 71.2 & 82.6 & - & - & - & - \\
        \hline
        $\text{IODINE}^\S$ & 30.9 & 44.6 & 54.4 & 67.0 & 51.7 & 67.3 & 19.9 & 32.4 & 7.3 & 12.8 \\
        $\text{Slot-Attn.}^\S$ & 35.6 & 51.5 & 38.6 & 55.3 & 41.3 & 58.3 & 83.6 & 90.7 & 7.3 & 13.5 \\
        Ours & \textbf{56.4} & \textbf{70.9} & \textbf{71.7} & \textbf{83.2} & \textbf{72.4} & \textbf{83.7} & \textbf{84.7} & \textbf{91.5} & \textbf{78.8} & \textbf{87.5} \\
        \bottomrule
    \end{tabular}
    \caption{\textbf{Foreground extraction results on training data measured in IoU and Dice.} Higher is better in all scores. *Results of W-Net and OneGAN are provided by \citet{benny2020onegan}. Of note, results of these two models on \ac{dogs} and \ac{cars} datasets may \textbf{not} be directly comparable to other listed methods, as the data used for training and evaluation could be different. We include these results as a rough reference since no official implementation or pretrained model are publicly available. \S ~ indicates unfair baseline results obtained using extra ground-truth information, \ie, we choose the best-matching scores from the permutation of foreground and background masks. \dag OneGAN is a strong \textbf{weakly supervised} baseline, which requires clean background images to provide additional supervision. We include this model as a potential upper bound of the fully unsupervised methods.}
    \label{tab:ext_results}
\end{table}

\paragraph{Single object in the wild}

In the first group of datasets, there is typically a single object in the foreground, varying in shapes, texture, and lighting conditions. Unsupervised foreground extraction on these datasets requires much more sophisticated visual cues than colors and shapes. \ac{birds} dataset consists of 11,788 images of 200 classes of birds annotated with high-quality segmentation masks, \ac{dogs} dataset consists of 20,580 images of 120 classes annotated with bounding boxes, and \ac{cars} dataset consists of 16,185 images of 196 classes. The latter two datasets are primarily made for fine-grained categorization. To evaluate foreground extraction, we follow the practice in \citet{benny2020onegan}, and approximate ground-truth masks for the images with Mask R-CNN~\cite{he2017mask}, pre-trained on the MS COCO~\cite{lin2014microsoft} dataset with a ResNet-101~\cite{he2016deep} backend. The pre-trained model is acquired from the detectron2~\cite{wu2019detectron2} toolkit. This results in 5,024 dog images and 12,322 car images with a clear foreground-background setup and corresponding masks. 

\begin{figure}[t!]
    \centering
    \includegraphics[width=\linewidth]{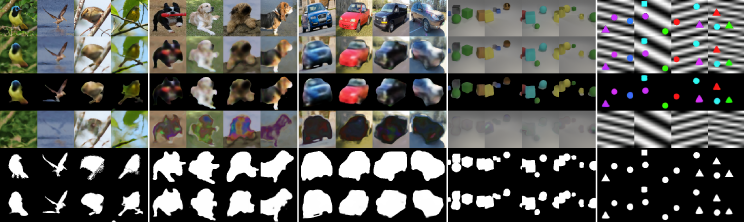}
    \caption{\textbf{Foreground extraction results for each dataset}; zoom in for better visibility. From top to bottom: (i) observed images, (ii) generated images, (iii) masked generated foregrounds, (iv) generated backgrounds, (v) ground-truth foreground masks, and (vi) inferred foreground masks. More samples and results of baselines can be found in the supplementary material.}
    \label{fig:viz}
\end{figure}

On datasets featuring a single foreground object, we use the 2-slot version of IODINE and Slot-attention. Since ReDO, IODINE, and Slot-Attention do not distinguish foreground and background in output regions, we choose the best-matching scores from the permutation of foreground and background masks as in~\cite{chen2019unsupervised}. We observe that the proposed method and Grabcut are the only two methods that provide explicit identification of foreground objects and background regions. While the Grabcut algorithm actually requires a predefined bounding box as input that specifies the foreground region, our method, thanks to the learned pixel re-assignment (see \cref{sec:model}), can achieve this in a fully unsupervised manner. Results in \cref{tab:ext_results} show that our method outperforms all the unsupervised baselines by a large margin, exhibiting comparable performance even to the weakly supervised baseline that requires additional background information as inputs~\cite{benny2020onegan}. We provide samples of foreground extraction results as well as generated background and foreground regions in \cref{fig:viz}. Note that our final goal is not to synthesize appealing images but to learn foreground extractors in a fully unsupervised manner. As the limitation of our method, \ac{drc} generates foreground and background regions less realistic than those generated by state-of-the-art GANs, which hints a possible direction for future work. More detailed discussions of the limitation can be found in supplementary material.

\paragraph{Multi-object scenes}

The second group of datasets contains images with possibly simpler foreground objects but more challenging scene configurations or background parts. Visual scenes in the CLEVR6 dataset contain various objects and often with partial occlusions and truncations.
Following the evaluation protocol in IODINE and Slot-attention, we use the first 70K samples from CLEVR~\cite{johnson2017clevr} and filter the samples for scenes with at most 6 objects for training and evaluation, \ie, CLEVR6. The \ac{tmds} dataset is a variant of Multi-dSprites~\cite{dsprites17} but has strongly confounding backgrounds borrowed from Textured MNIST~\cite{greff2016tagger}. We generate 20K samples for the experiments. Similar to \citet{greff2019multi} and \citet{locatello2020object}, we evaluate on a subset containing 1K samples for testing. 
Note that IODINE and Slot-attention are designed for segmenting complex multi-object scenes using slot-based object representations. Ideally, the output of these models consists of masks for each individual object, while the background is viewed as a virtual ``object'' as well. In practice, however, it is possible that the model distributes the background over all the slots as mentioned in \citet{locatello2020object}. We therefore propose two corresponding approaches (see the supplementary material for more details) to convert the output object masks into a foreground-background partition and report the best results of these two options for IODINE and Slot-attention in \cref{tab:ext_results}. 

On the CLEVR6 dataset, we use the publicly available pretrained model for IODINE, which achieves a reasonable ARI (excluding background pixels) of 94.4 on the testing data, close to the testing results in \citet{greff2019multi}. We observe that IODINE distributes the background over all the slots for some of the testing samples, resulting in much lower IoU and Dice scores. We re-train the Slot-attention model using the official implementation on CLEVR6, as no pretrained model is publicly available. The re-trained model achieves a foreground ARI of 98.0 on the 1K testing samples, which we consider as a sign of valid re-implementation. Results in \cref{tab:ext_results} demonstrate that the proposed method can effectively process images of challenging multi-object scenes. To be specific, our method demonstrates competitive performance on the CLEVR6 dataset compared with the SOTA object discovery method. Moreover, as shown empirically in \cref{fig:viz}, the proposed method can handle the strongly confounding background introduced in \citet{greff2016tagger}, whereas previous methods are distracted by the background and mostly fail to capture the foreground objects.

\begin{wraptable}{r}{0.4\linewidth}
    \vspace{-15pt}
    \centering
    \small
    \begin{tabular}{l|cc}
        \toprule
        \multicolumn{1}{c|}{Model} & 
        \multicolumn{1}{c}{IoU} &
        \multicolumn{1}{c}{Dice} \\
        \hline
        $\text{amortized inference}^{*}$ & - & - \\
        w/o pix. re-assign. & 21.8 & 35.3 \\
        \hline
        w/o pseudo label & 48.7  & 64.2 \\
        w/o TV-norm reg. & 53.0  & 68.1 \\
        w/o ortho. reg. & 54.3  & 69.2 \\
        \hline
        $\text{short-run chain}^\dag$ & 52.5 & 67.7 \\
        \hline
        $\text{Full model}$ & 56.4 & 70.9 \\
        \bottomrule
    \end{tabular}
    \caption{\textbf{Ablation study on Birds.} *We replace the \ac{lebm} with encoders to perform amortized inference for the latent variables $\z$ within a variational framework as in VAE~\cite{kingma2013auto}. \dag We explore the possibility of using short-run MCMC~\cite{nijkamp2019learning} instead of persistent chain sampling.}
    \label{tab:ablation}
\end{wraptable}

\subsection{Ablation Study}\label{sec:ablation_study}

We provide ablation studies on the \ac{birds} dataset to inspect the contribution of each proposed component in our model. As shown in \cref{tab:ablation}, we observe that replacing the \ac{lebm}s in the foreground and background models with amortized inference networks significantly harms the performance of the proposed method. In particular, the modified model fails to generate any meaningful results (marked as - in \cref{tab:ablation}). We conjecture that \ac{lebm} benefits from the low-dimensionality of the latent space~\cite{pang2020learning} and therefore enjoys more efficient learning.
However, the inference networks need to learn an extra mapping from the high-dimensional image space to the latent space and require more elaborate architecture and tuning for convergence. Furthermore, we observe that the model that does not learn pixel re-assignment for background can still generate meaningful images but only vaguely captures masks for foreground extraction.

\subsection{Generalizable Foreground Extraction}\label{sec:gen_fe}

\paragraph{Extracting novel foreground objects from training categories}

We show results on generalizing to novel objects from the training classes. To evaluate our method, we split the \ac{birds} dataset following \citet{chen2019unsupervised}, resulting in 10K training images and 1K testing images. On \ac{dogs} and \ac{cars} datasets, we split the dataset based on the original train-test split~\cite{khosla2011novel,krause20133d}. This split gives 3,286 dog images and 6,218 car images for training, and 1,738 dog images and 6,104 car images for testing, respectively. As summarized in \cref{tab:gen_from_tr}, our method shows superior performances compared with baselines; the performance gap between training and testing is constantly small over all datasets.

\begin{table}[!htbp]
    \centering
    \begin{tabular}{lcccccc}
        \toprule
        \multicolumn{1}{c}{ } & 
        \multicolumn{2}{c}{\ac{birds}} &
        \multicolumn{2}{c}{\ac{dogs}} &
        \multicolumn{2}{c}{\ac{cars}} \\
        \cmidrule(r){2-3} \cmidrule(r){4-5} \cmidrule(r){6-7}
        
        \multicolumn{1}{c}{Model} & 
        \multicolumn{1}{c}{IoU} & 
        \multicolumn{1}{c}{Dice} &
        \multicolumn{1}{c}{IoU} & 
        \multicolumn{1}{c}{Dice} & 
        \multicolumn{1}{c}{IoU} & 
        \multicolumn{1}{c}{Dice} \\
        \cmidrule(r){2-2} \cmidrule(r){3-3} \cmidrule(r){4-4}
        \cmidrule(r){5-5} \cmidrule(r){6-6} \cmidrule(r){7-7}
        
        \multicolumn{1}{c}{} & 
        \multicolumn{1}{c}{Tr.|Te.} & 
        \multicolumn{1}{c}{Tr.|Te.} & 
        \multicolumn{1}{c}{Tr.|Te.} & 
        \multicolumn{1}{c}{Tr.|Te.} & 
        \multicolumn{1}{c}{Tr.|Te.} & 
        \multicolumn{1}{c}{Tr.|Te.} \\
        \hline
        $\text{GrabCut}^{*}$ & 30.2|30.3 & 42.7|42.8 & 58.3|57.9 & 70.8|70.5 & 60.9|61.6 & 72.7|73.5 \\
        ReDO & 46.8|47.1 & 61.4|61.7 & 54.3|52.8 & 69.2|67.9 & 52.6|52.5 & 68.7|68.6\\
        Ours & \textbf{54.8|54.6} & \textbf{69.5|69.4} & \textbf{71.6|72.3} & \textbf{83.2|83.6} & \textbf{71.9|70.8} & \textbf{83.3|82.5} \\
        \bottomrule
    \end{tabular}
    \caption{\textbf{Performance of \ac{drc} on training and held-out testing data.} *Note that GrabCut is a deterministic method that does not require training. We report the results of GrabCut~\cite{rother2004grabcut} on these splits only for reference. Tr. indicates the performance on training data, and Te. indicates the performance on testing data.}
    \label{tab:gen_from_tr}
\end{table}

\paragraph{Extracting novel foreground objects from unseen categories}

\begin{wraptable}{r}{0.435\linewidth}
\small{
    \centering
    \setlength\tabcolsep{1.2pt}
    \begin{tabular}{cccccccc}
        \toprule
        \multicolumn{1}{c}{\multirow{2}{*}{Test}} & 
        \multicolumn{1}{c}{\multirow{2}{*}{Train}} & 
        \multicolumn{2}{c}{GrabCut} &
        \multicolumn{2}{c}{ReDO} &
        \multicolumn{2}{c}{Ours} \\
        \cmidrule(r){3-4} \cmidrule(r){5-6} \cmidrule(r){7-8}
        
        \multicolumn{2}{c}{} & 
        \multicolumn{1}{c}{\small{IoU}} & 
        \multicolumn{1}{c}{\small{Dice}} &
        \multicolumn{1}{c}{\small{IoU}} & 
        \multicolumn{1}{c}{\small{Dice}} &
        \multicolumn{1}{c}{\small{IoU}} & 
        \multicolumn{1}{c}{\small{Dice}} \\
        
        \hline
        
        {} & \ac{birds}* & {} & {}
               & 47.1 & 61.7
               & 54.6 & 69.4 \\
        \cmidrule(r){2-2} \cmidrule(r){5-8}
        \ac{birds} & \ac{dogs} & 30.3 & 42.8
                      & 22.2 & 35.3
                      & \textbf{41.3} & \textbf{57.4} \\
        \cmidrule(r){2-2} \cmidrule(r){5-8}
        {} & \ac{cars}  & {} & {} 
                & 16.4 & 27.7 
                & 39.2 & 55.3 \\
        \hline
        
        {} & \ac{dogs}* & {} & {}
               & 52.8 & 67.9
               & 72.3 & 83.6 \\
        \cmidrule(r){2-2} \cmidrule(r){5-8}
        \ac{dogs} & \ac{cars} & 57.9 & 70.5
               & 44.5 & 61.2
               & \textbf{67.8} & \textbf{80.4} \\
        \cmidrule(r){2-2} \cmidrule(r){5-8}
        {} & \ac{birds}  & {} & {} 
                 & 44.0 & 60.3
                 & 53.6 & 69.1 \\
        \hline                  
        
        {} & \ac{cars}* & {} & {}
               & 52.5 & 68.6
               & 70.8 & 82.5 \\
        \cmidrule(r){2-2} \cmidrule(r){5-8}
        \ac{cars} & \ac{dogs} & 61.6 & 73.5
               & 51.6 & 67.1
               & \textbf{68.6} & \textbf{81.0} \\
        \cmidrule(r){2-2} \cmidrule(r){5-8}
        {} & \ac{birds}  & {} & {} 
                 & 41.8 & 58.6
                 & 45.1 & 61.7 \\
        \bottomrule
    \end{tabular}
    }
    \caption{\textbf{Performance of \ac{drc} on unseen testing categories.} *We include the testing results of models trained with data from the same categories for reference.}
    \label{tab:gen_novel}
\end{wraptable}

To investigate how well our method generalizes to categories unseen during training, we evaluate the models trained in real-world single object datasets on the held-out testing data from different categories. We use the same training and testing splits on these datasets as in the previous experiments. \cref{tab:gen_novel} shows that our method outperforms the baselines on the \ac{birds} dataset when the model has trained on \ac{dogs} or \ac{cars} dataset, which have quite different distributions in foreground object shapes. Competitors like ReDO also exhibit such out-of-distribution generalization but only to a limited extent. Similar results are observed when using \ac{dogs} or \ac{cars} as the testing set. We can see that when the model is trained on \ac{dogs} and evaluated on \ac{cars} or vice versa, it still maintains comparable performances \wrt those are trained\&tested on the same class. We hypothesize that these two datasets have similar distributions in foreground objects and background regions. In the light of this, we can further entail that the distribution of \ac{dogs} is most similar to that of \ac{cars} and less similar to that of \ac{birds} according to the results, which is consistent with our intuitive observation of the data. We provide a preliminary analysis of the statistics of these datasets in the supplementary material.

\section{Conclusion}

We have presented the Deep Region Competition, an \ac{em}-based fully unsupervised foreground extraction algorithm fueled by energy-based prior and generative image modeling. We propose learned pixel re-assignment as an inductive bias to capture the background regularities. Experiments demonstrate that \ac{drc} exhibits more competitive performances on complex real-world data and challenging multi-object scenes. We show empirically that learned pixel re-assignment helps to provide explicit identification for foreground and background regions. Moreover, we find that \ac{drc} can potentially generalize to novel foreground objects even from categories unseen during training. We hope our work will inspire future research along this challenging but rewarding research direction.

\section*{Acknowledgements}
The work was supported by NSF DMS-2015577, ONR MURI project N00014-16-1-2007, and DARPA XAI project N66001-17-2-4029. We would like to thank Bo Pang from the UCLA Statistics Department for his insights on the latent-space energy-based model and four anonymous reviewers (especially Reviewer RcgA) for their constructive comments.

\small
\bibliographystyle{unsrtnat}
\bibliography{reference}

\newpage
\section*{Checklist}


\begin{enumerate}

\item For all authors...
\begin{enumerate}
  \item Do the main claims made in the abstract and introduction accurately reflect the paper's contributions and scope?
    \answerYes{See \cref{sec:results}, \cref{sec:ablation_study} and \cref{sec:gen_fe}.}
  \item Did you describe the limitations of your work?
    \answerYes{See \cref{sec:results}.}
  \item Did you discuss any potential negative societal impacts of your work?
    \answerNA{}
  \item Have you read the ethics review guidelines and ensured that your paper conforms to them?
    \answerYes{}
\end{enumerate}

\item If you are including theoretical results...
\begin{enumerate}
  \item Did you state the full set of assumptions of all theoretical results?
    \answerNA{}
	\item Did you include complete proofs of all theoretical results?
    \answerNA{}
\end{enumerate}

\item If you ran experiments...
\begin{enumerate}
  \item Did you include the code, data, and instructions needed to reproduce the main experimental results (either in the supplemental material or as a URL)?
    \answerYes{We provide pytorch-style codes and detailed instructions of how to reproduce the main results in the supplemental material.}
  \item Did you specify all the training details (e.g., data splits, hyperparameters, how they were chosen)?
    \answerYes{See \cref{sec:results}, \cref{sec:ablation_study}, \cref{sec:gen_fe} and supplemental material.}
	\item Did you report error bars (e.g., with respect to the random seed after running experiments multiple times)?
    \answerNA{}
	\item Did you include the total amount of compute and the type of resources used (e.g., type of GPUs, internal cluster, or cloud provider)?
    \answerYes{See supplemental material.}
\end{enumerate}

\item If you are using existing assets (e.g., code, data, models) or curating/releasing new assets...
\begin{enumerate}
  \item If your work uses existing assets, did you cite the creators?
    \answerYes{}
  \item Did you mention the license of the assets?
    \answerNA{}
  \item Did you include any new assets either in the supplemental material or as a URL?
    \answerNA{}
  \item Did you discuss whether and how consent was obtained from people whose data you're using/curating?
    \answerYes{See \cref{sec:results}, \cref{sec:ablation_study} and \cref{sec:gen_fe}.}
  \item Did you discuss whether the data you are using/curating contains personally identifiable information or offensive content?
    \answerNA{}
\end{enumerate}

\item If you used crowdsourcing or conducted research with human subjects...
\begin{enumerate}
  \item Did you include the full text of instructions given to participants and screenshots, if applicable?
    \answerNA{}
  \item Did you describe any potential participant risks, with links to Institutional Review Board (IRB) approvals, if applicable?
    \answerNA{}
  \item Did you include the estimated hourly wage paid to participants and the total amount spent on participant compensation?
    \answerNA{}
\end{enumerate}
\end{enumerate}

\clearpage
\renewcommand\thefigure{S\arabic{figure}}
\setcounter{figure}{0}
\renewcommand\thetable{S\arabic{table}}
\setcounter{table}{0}
\pagenumbering{arabic}
\renewcommand*{\thepage}{S\arabic{page}}
\appendix

\section{Dataset Details}
\subsection{Caltech-UCSD Birds-200-2011}
\ac{birds} dataset consists of 11,788 images of 200 classes of birds annotated with high-quality segmentation masks. Each image is further annotated with 15 part locations, 312 binary attributes, and 1 bounding box. We use the provided bounding box to extract a center square from the image, and scale it to $128 \times 128$ pixels. Each scene contains exactly one foreground object. 

\subsection{Stanford Dogs}
\begin{figure}[!htbp]
    \centering
    \includegraphics[width=\textwidth]{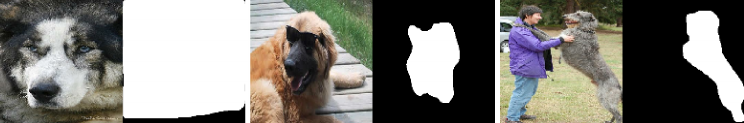}
    \caption{Examples of excluded images. From left to right: (i) image with a foreground object that occupied too much space, (ii) image with a low-quality mask, and (iii) image with significant foreground distractors.}
    \label{fig:excluded_dogs}
\end{figure}

\ac{dogs} dataset consists of 20,580 images of 120 classes annotated with bounding boxes. We first use the provided bounding box to extract the center square, and then scale it to $128 \times 128$ pixels. As stated in the paper, we approximate ground-truth masks for the pre-processed images with Mask R-CNN~\cite{he2017mask}, pre-trained on the MS COCO~\cite{lin2014microsoft} dataset with a ResNet-101~\cite{he2016deep} backend. The pretrained model is acquired from the detectron2~\cite{wu2019detectron2} toolkit. We exclude the images where no dog is detected. We then manually exclude those images where the foreground object has occupied more than $\sim90\%$ of the image, those with poor masks, and those with significant foreground distractors such as humans (see \cref{fig:excluded_dogs}). The filtering strategy results in 5,024 images with a clear foreground-background setup and high-quality mask.

\subsection{Stanford Cars}
\ac{cars} dataset consists of 16,185 images of 196 classes annotated with bounding boxes. Though also being primarily designed for fine-grained categorization, it has a much clearer foreground-background setup compared with the \ac{dogs} dataset. We employ a similar process as used for \ac{dogs} dataset to approximate the ground-truth masks, and only exclude those images where cars are not properly detected. It finally produces 12,322 images for our experiments.

\subsection{CLEVR6}
CLEVR6 dataset is a subset of the original CLEVR dataset~\cite{johnson2017clevr} with masks, generated by~\citet{greff2019multi}. We follow the evaluation protocol adopted by IODINE~\cite{greff2019multi} and Slot-attention~\cite{locatello2020object}, which takes the first 70K samples from CLEVR. These samples are then filtered to only include scenes with at most 6 objects. Additionally, we perform a center square crop of $192 \times 192$ from the original $240 \times 320$ image, and scale it to $128 \times 128$ pixels. The resulting CLEVR6 dataset contains 3-6 foreground objects that could be with partial occlusion and truncation in each visual scene.

\subsection{Textured Multi-dSprites}
\begin{figure}[!htbp]
    \centering
    \includegraphics[width=\linewidth]{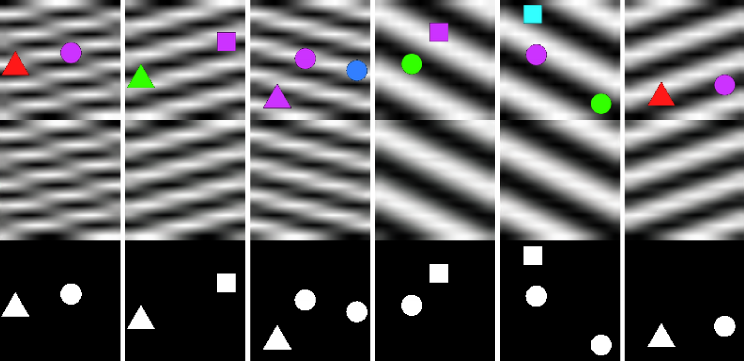}
    \caption{Samples from \ac{tmds}. From top to bottom: (1) observed images, (ii) background textures, and (iii) ground-truth masks.}
    \label{fig:example_textured}
\end{figure}

\ac{tmds} dataset, which is based on the dSprites dataset~\cite{dsprites17} and Textured MNIST~\cite{greff2016tagger}, consists of 20,000 images with a resolution of $128 \times 128$. Each image contains 2-3 random sprites, which vary in terms of shape (square, circle, or triangle), color (uniform saturated colors), and position (continuous). The background regions are borrowed from Textured MNIST dataset~\cite{greff2016tagger}. The textures for the background are randomly shifted samples from a bank of 20 sinusoidal textures with different frequencies and orientations. We adopt a simpler foreground setting compared with the vanilla Multi-dSprites dataset used in~\cite{greff2019multi}, \ie, the foreground objects are not occluded as the dataset is designed to emphasize the background part. Some samples are presented in \cref{fig:example_textured}.

\newpage
\section{Details on Models and Hyperparameters}
\label{sec:appendix_model_and_arch}
\paragraph{Architecture}
As mentioned in the paper, we use the same overall architecture for different datasets (while the size of latent variables may vary). The details for the generators and LEBMs are summarized in the \cref{tab:hdim} and \cref{tab:arch}.
\begin{table}[!htbp]
    \centering
    \begin{tabular}{l|ccc}
    \toprule
    Dataset & Foreground & Background & Pixel Re-assignment \\
    \cmidrule(r){1-1} \cmidrule(r){2-2} \cmidrule(r){3-3}
    \cmidrule(r){4-4}
    \ac{birds} & 256 & 256 & 512 \\
    \ac{dogs} & 256 & 256 & 512 \\
    \ac{cars} & 256 & 192 & 512 \\
    CLEVR6 & 256 & 2 & 256 \\
    \ac{tmds} & 256 & 4 & 1024 \\
    \bottomrule
    \end{tabular}
    \caption{Dimension of latent variables on each dataset.}
    \label{tab:hdim}
\end{table}

\begin{table}[!htbp]
    \centering
    \begin{tabular}{ccc}
    \toprule
    Layers & In-Out size & Comment \\
    \cmidrule(r){1-1} \cmidrule(r){2-2} \cmidrule(r){3-3}
    
    \multicolumn{3}{c}{LEBM for Foreground/Background Models} \\
    \hline
    Input: $\z$ & $D^{*}$ & {} \\
    Linear, LReLU & 200 & {} \\
    Linear, LReLU & 200 & {} \\
    Linear & $K^\dag$ & {} \\
    
    \hline
    \multicolumn{3}{c}{LEBM for Pixel Re-assignment Model} \\
    \hline
    Input: $\z$ & $D^{*}$ & {} \\
    Linear, LReLU & 200 & {} \\
    Linear, LReLU & 200 & {} \\
    Linear, LReLU & 200 & {} \\
    Linear & 1 & {} \\
    
    \hline
    \multicolumn{3}{c}{Generator for Foreground/Background Model and Re-assignment Model} \\
    \hline
    Input: $\z$ & $D^{*}$ & {} \\
    Linear, LReLU & $4 \times 4 \times 128$ & reshaped output \\
    UpConv3x3Norm, LReLU & $8 \times 8 \times 1024$ & stride 1 \& padding 1 \\
    UpConv3x3Norm, LReLU & $16 \times 16 \times 512$ & stride 1 \& padding 1 \\
    UpConv3x3Norm, LReLU & $32 \times 32 \times 256$ & stride 1 \& padding 1 \\
    UpConv3x3Norm, LReLU & $64 \times 64 \times 128$ & stride 1 \& padding 1 \\
    UpConv3x3Norm, LReLU & $128 \times 128 \times 64$ & stride 1 \& padding 1 \\
    Conv3x3 & \tabincell{c}{$128 \times 128 \times (3 + 1)$ \\
                            $128 \times 128 \times 2$} 
            & \tabincell{c}{RGB \& Mask \\ Re-assignment grid} \\
    
    \hline
    \multicolumn{3}{c}{Auxiliary classifier for Foreground/Background Model} \\
    \hline
    Input: $\x$ & $128 \times 128 \times 3$ & generated image \\
    Conv4x4Norm, LReLU & $64 \times 64 \times 64$ & stride 2 \& padding 1 \\
    Conv4x4Norm, LReLU & $32 \times 32 \times 128$ & stride 2 \& padding 1 \\
    Conv4x4Norm, LReLU & $16 \times 16 \times 256$ & stride 2 \& padding 1 \\
    Conv4x4Norm, LReLU & $8 \times 8 \times 512$ & stride 2 \& padding 1 \\
    Conv4x4Norm, LReLU & $4 \times 4 \times 1024$ & stride 2 \& padding 1 \\
    Conv4x4 & $1 \times 1 \times K^\dag$ & {} \\
    
    \bottomrule
    \end{tabular}
    \caption{Architecture of the generators, LEBMs and auxiliary classifiers (see \cref{sec:reg}). UpConv3x3Norm denotes a Upsampling-Convolutional-InstanceNorm layer with a convolution kernel size of 3. Similarly, Conv4x4Norm denotes a Convolutional-InstanceNorm layer with a kernel size of 4. LReLU denotes the Leaky-ReLU activation function. The leak factor for LReLU is $0.2$ in LEBMs and auxiliary classifiers, and $0.01$ in generators. *$D$ represents the dimensions of the latent variables for different datasets; see \cref{tab:hdim}. \dag $K$ represents the pre-specified category number for latent variables. We use 200 for both the foreground and background LEBMs on real-world datasets, and 30 and 10 in the foreground and background LEBMs on multi-object datasets respectively.}
    \label{tab:arch}
\end{table}

\paragraph{Hyperparameters and Training Details}
For the Langevin dynamics sampling~\cite{welling2011bayesian}, we use $K_0$ and $K_1$ to denote the number of prior and posterior sampling steps with step sizes $s_0$ and $s_1$ respectively. Our hyperparameter choices are: $K_0=60, K_1=40, s_0=0.4$ and $s_1=0.1$. These are identical across different datasets. 
During testing, we set the posterior sampling steps to 300 for \ac{dogs} and \ac{cars}, and 2.5K, 5K and 5K for \ac{birds}, CLEVR6 and \ac{tmds} respectively. The parameters of the generators and LEBMs are initialized with orthogonal initialization~\cite{saxe2013exact}. The gain is set to $1.0$ for all the models. We use the ADAM optimizer \cite{kingma2014adam} with $\beta_1=0.5$ and $\beta_2=0.999$. Generators are trained with a constant learning rate of $0.0001$, and LEBMs with $0.00002$. We run experiments on a single V100 GPU with 16GB of RAM and with a batch size
of 48. We set the maximum training iterations to 10K and run for at most 48hrs for each dataset.

\newpage
\section{Details on Learning Objective and Regularization}
\subsection{Learning Objective}
\paragraph{Derivation of Surrogate Learning Objective}
$\mathcal{J}(\theta)=\mathbf{E}_{\w \sim p_\beta(\w | \x, \z)}\left[\mathcal{L}(\theta)\right]$ is the conditional expectation of $\w$, 
\begin{equation}
    \begin{aligned}
    \mathcal{J}(\theta) &= \mathbf{E}_{\w \sim p_\beta(\w | \x, \z)}
                            \left[
                                \mathcal{L}(\theta)
                            \right] \\
             &= \log p_\alpha(\z) + \mathbf{E}\left[
                    \sum\limits_{i=1}^D\sum\limits_{k=1}^2 w_{ik} 
                    \left(
                        \log \pi_{ik} + \log p_{\beta_k}(\x_i | \z_k)
                    \right)
                \right] \\
            &= \log p_\alpha(\z) + 
                    \sum\limits_{i=1}^D\sum\limits_{k=1}^2 \mathbf{E}\left[ w_{ik} \right]
                    \left(
                        \log \pi_{ik} + \log p_{\beta_k}(\x_i | \z_k)
                    \right),
    \end{aligned}
\end{equation}
where $\mathbf{E}$ is the conditional expectation of $\w$. Recall that $w_{ik} \in \{0, 1\}$. The expectation becomes
\begin{equation}
    \begin{aligned}
    \mathbf{E}\left[ w_{ik} \right] &= 0 \times p(w_{ik} = 0|\x_i, \z) 
                                     + 1 \times p(w_{ik} = 1|\x_i, \z) \\
                                    &= \gamma_{ik},
    \end{aligned}
\end{equation}
which is the posterior responsibility of $w_{ik}$. We can further decompose $\mathcal{J}(\theta)$ into
\begin{equation}
    \mathcal{J}(\theta) = 
            \underbrace{
                \log p_\alpha(\z)
            }_{\text{objective for \ac{lebm}}}
          + \underbrace{
                \sum\limits_{i=1}^D\sum\limits_{k=1}^2 \gamma_{ik}\log \pi_{ik}
            }_{\text{foreground-background partitioning}}
          + \underbrace{
                \sum\limits_{i=1}^D\sum\limits_{k=1}^2 \gamma_{ik}\log p_{\beta_k}(\x_i | \z_k)
            }_{\text{objective for image generation}},
\end{equation}
as mentioned in the paper.

\paragraph{Understanding the Optimization Process} Note that the surrogate learning objective is an expectation w.r.t $\z$,
\begin{equation}
    \max\limits_{\theta}~ \mathbf{E}_{\z \sim p_\theta(\z|\x)} 
            \left[
                \mathcal{J}(\theta) 
            \right], ~
        \text{s.t.} ~\forall i, \sum\limits_{k=1}^2 \pi_{ik} = 1,
\end{equation}
which is generally intractable to calculate. We therefore need to approximate the expectation by sampling from the distributions, and calculating the Monte Carlo average. In practice, this can be done by gradient-based MCMC sampling method, such as Langevin Dynamics~\cite{welling2011bayesian}. 

Given $\x$, we have $p_\theta(\z | \x) \propto p_\beta(\x | \z) p_\alpha(\z)$. Note that
\begin{equation}
    \begin{aligned}
    \nabla_\z \log p_\beta(\x | \z) &= \frac{1}{p_\beta(\x | \z)} \nabla_\z p_\beta(\x | \z) \\
        &= \int_\w p_\beta(\w | \x, \z) \nabla_\z \log p_\beta(\x, \w | \z) d\w \\
        &= \mathbf{E}_{\w \sim p_\beta(\w | \x, \z)} \left[
                \nabla_\z \log p_\beta(\x, \w | \z)
           \right].
    \end{aligned}
\end{equation}

Therefore, the log-likelihood of surrogate target distribution for the Langevin dynamics at the $t$-th step is
\begin{equation}
    \begin{aligned}
    \log \tilde{Q}(\z_{t}) &= \log p_\alpha(\z_{t}) 
             + \mathbf{E}_{\w \sim p_\beta(\w| \x, \z_{t})} \left[
                \sum\limits_{i=1}^D\sum\limits_{k=1}^2 w_{ik} 
                    \left(
                        \log \pi_{ik} + \log p_{\beta_k}(\x_i | \z_{k, t})
                    \right)
             \right] \\
            &= \log p_\alpha(\z_{t}) +  
                \sum\limits_{i=1}^D\sum\limits_{k=1}^2 
                \gamma_{ik, t}
                    \left(
                        \log \pi_{ik} + \log p_{\beta_k}(\x_i | \z_{k, t})
                    \right), 
    \end{aligned}
\end{equation}
which has the same form as $\mathcal{J}(\theta)$. However, instead of updating parameters $\theta$, Langevin dynamics updates the latent variables $\z$ with the calculated gradients.

The two-step learning process of the DRC models can be understood as follows: (1) in the first step, the algorithm optimizes $\mathcal{J}$ by updating latent variables $\z$, where the posterior responsibility $\gamma_{ik}$ inferred at each step serves to gradually disentangle the foreground and background components, and (2) in the second step, the updated $\z$ is fed again into the models to generate the observation $\x$, where the algorithm optimizes $\mathcal{J}$ by updating the model parameters $\theta$.

It is worth mentioning that learning LEBMs requires an extra sampling step~\cite{pang2020learning}, as the gradients are given by the following
\begin{equation}
    \delta_\alpha (\x) = \E_{p_\theta(\z|\x)} \left[\nabla_\alpha f_\alpha(\z) \right]
                       - \E_{p_\alpha(\z)} \left[\nabla_\alpha f_\alpha(\z) \right],
\end{equation}
where the second terms should be computed by sampling with $p_\alpha(\z)$. We term this as \textit{prior sampling} in the main paper.

\paragraph{Further Details on the Loss Functions}
For the generative models $p_{\beta_k}(\x | \z_k),~k=1,2$, we assume that $\x = g_{\beta_k}(\z_k)+ \epsilon$, where $g_{\beta_k}(\z_k),~k=1,2$ are the generator networks for foreground and background regions, and $\epsilon$ is random noise sampled from a zero-mean Gaussian or Laplace distribution. Assuming a global fixed variance $\sigma^2$ for Gaussian, we have $\log p_{\beta_k}(\x | \z_k) = - \frac{1}{2\sigma^2}\|g_{\beta_k}(\z_k) - \x \|^2 + C,~k=1,2$, where $C$ is a constant unrelated to $\beta_k$ and $\z_k$. Similarly for Laplace distribution, we have $\log p_{\beta_k}(\x | \z_k) = - \frac{1}{\lambda}|g_{\beta_k}(\z_k) - \x | + C,~k=1,2$. These two log-likelihoods correspond to the MSE loss and L1 loss commonly used for image reconstruction, respectively.

\subsection{Regularization}\label{sec:reg}
\paragraph{Pseudo Label Learning}
As mentioned in the paper, we exploit the symbolic vector $\y$ emitted by the LEBM for additional regularization. Let the target distribution of $\y_k$ be $P_k$ given by $p_{\alpha_k}(\y | \z_k),~k=1,2$, which represents the distribution of symbolic vector for foreground and background regions respectively. We can optimize the following objective as a regularization to our original learning objective:
\begin{equation}
    \max\limits_{\beta, \tau}~\mathcal{L}_{\text{pseudo-label}} = \sum\limits_{k=1}^2 H(P_k, Q_k),
\end{equation}

\begin{equation}
    H(P_k, Q_k) = - \langle 
                        p_{\alpha_k}(\y | \z_k), 
                        \log q_{\tau_k}(\y | g_{\beta_k}(\z_k)) 
                    \rangle,~k=1, 2,
\end{equation}
where $q_{\tau_k},~k=1,2$ represents the jointly trained auxiliary classifier network (see \cref{sec:appendix_model_and_arch} for architecture details) for foreground and background. $g_{\beta_k}(\z_k),~k=1,2$ represents the output of generator network. We set the weight of this regularization term to $0.1$ for all the models.

\paragraph{\ac{tvn}}
Total Variation norm~\cite{rudin1992nonlinear} is commonly used for image denoising, and has been extended as an effective technique for in-painting. We use \ac{tvn} as a regularization for learning the background generator.
\begin{equation}
    \min_{\beta_2}~\mathcal{L}_{\text{TV-norm}} = 
            \sum\limits_{h, w} \left(
             |\frac{\partial g_{\beta_2}(\z_2)}{\partial x}(h, w)| 
           + |\frac{\partial g_{\beta_2}(\z_2)}{\partial y}(h, w)|
           \right),
\end{equation}
where $\partial_x g_{\beta_2}(\z_2)(h, w)$ and $\partial_y g_{\beta_2}(\z_2)(h, w)$ represent the horizontal and vertical image gradients at the pixel coordinate $(h, w)$ respectively. We set the weight of this regularization term to $0.01$ for all the models.

\paragraph{Orthogonal Regularization}
We use orthogonal regularization~\cite{brock2016neural} for the convolutional layers only. Let $\mathbf{W}$ be the flattened kernel weights of the convolutional layers, \ie, the size of $\mathbf{W}$ is $C \times K$ where $C$ is the output channel number. The orthogonal regularization is calculated according to
\begin{equation}
    \min_{\beta}~\mathcal{L}_{\text{orthogonal-reg}} = \|\mathbf{W}\mathbf{W}^T \odot (\mathbf{1} - \mathbf{I}) \|_{F},
\end{equation}
where $\odot$ is the Hadamard product. $\mathbf{I}$ denotes the identity matrix, and $\mathbf{1}$ denotes the matrix filled with ones. We set the weight of this regularization term to $0.1$ for \ac{birds} models, and $1.0$ for the rest of the models.

\newpage
\section{Pytorch-style Code}
We provide pytorch-style code to illustrate how the learning and inference in our model work.

\paragraph{Forward Pass} In the forward pass, the model takes latent variables $\z$, generates foreground and background regions separately, and mixes them into the final image. Note that the pixel re-assignment is applied to both background image and mask. We finds it useful to feed the intermediate feature of background region into the generator for pixel re-assignment.

\begin{lstlisting}[language=Python, caption={Forward pass of the DRC model.}]
def forward(z):
    zf, zb, zs = z[:, :ZF_DIM], \
                 z[:, ZF_DIM:-ZS_DIM] \
                 z[:, -ZS_DIM:]
    
    ### generating foreground
    fg, fm = fg_net(zf)

    ### generating background
    bg, bm, bg_feat = bg_net(zb)
    shuffling_grid = sp_net(zs, bg_feat.detach())
    bg_shuf = F.grid_sample(bg, shuffling_grid)
    bm_shuf = F.grid_sample(bm, shuffling_grid)
        
    ### generating foreground masks 
    pi = torch.cat([fm_wp, bm_wp], dim=1).softmax(dim=1)
    pi_f, pi_b = pi[:,:1,...], pi[:,1:,...]

    ### mixing regions
    im_p = fg * pi_f + bg_wp * pi_b
    return im_p, fg, bg_shuf, pi_f, pi_b, bg
\end{lstlisting}

\newpage
\paragraph{Sampling Latent Variables} We employ Langevin Dynamics for sampling latent variables, which iteratively updates the sample with the gradient computed against the likelihood.
In the following code, \texttt{ebm\_net} stands for the LEBMs, which outputs the energy and the distribution paramters of the symbolic vector $\y$ for the foreground and background regions.
\begin{lstlisting}[language=Python, caption={Running Langevin Dynamics for prior and posterior sampling of the latent variables $\z$.}]
def sample_langevin_prior(z):
    ### langevin prior inference
    # only latent variables 'z' are updated
    for __ in range(infer_step_K0):
        z = z.requires_grad_(True)
        
        en, __, __ = ebm_net(z)
        e_log_lkhd = en.sum() + .5 * z.square().sum()
            
        d_ebm_z = torch.autograd.grad(e_log_lkhd, z)[0]
        z = z - 0.5 * (self.delta_0 ** 2) * d_ebm_z \
                    + self.delta_0 * torch.randn_like(z)
    return z

def sample_langevin_posterior(z, im_t):
    ### langevin posterior inference
    # only latent variables 'z' are updated
    for __ in range(infer_step):
        z = z.requires_grad_(True)
            
        im_p, fg, bg_shuf, pi_f, pi_b, bg = forward(z)

        ### log-lkhd for LEBMs
        en, __, __ = ebm_net(z)

        ### log-lkhd for generators
        log_pf = - F.l1_loss(fg, im_t, reduction='none')
                 / (2. * SIGMA ** 2)
        log_pb = - F.l1_loss(bg_shuf, im_t, reduction='none') 
                 / (2. * SIGMA ** 2)
            
        # posterior responsibility
        with torch.no_grad():
            ga_f = pi_f * log_pf.exp() / 
                  (pi_f * log_pf.exp() 
                 + pi_b * log_pb.exp() + 1e-8)
        # objective for image generation
        e_z_log_p = ga_f.detach() * \
                    ((pi_f + 1e-8).log() + log_pf) \
                  + (1. - ga_f.detach()) * \
                    ((pi_b + 1e-8).log() + log_pb)
        # regularization
        tv_norm = tv_loss(bg_shuf)

        j_log_lkhd = - e_z_log_p.sum() + tv_norm * .01 + \
                   + en.sum() + .5 * z.square().sum()

        d_j_z = torch.autograd.grad(j_log_lkhd, z)[0]
        z = z - 0.5 * (self.delta_1 ** 2) * d_j_z \
               + self.delta_1 * torch.randn_like(z)
        z = z.detach()
    return z
\end{lstlisting}

\newpage
\paragraph{Updating Model Parameters}
Given the sampled latent variables $\z$, we optimize the model parameters by minimizing the reconstruction error.
\begin{lstlisting}[language=Python, caption=Updating model parameters.]
def update_G(im_t, fg, bg_shuf, pi_f, pi_b, bg,
             zf_logits, zb_logits):
    ### optimizers for generator networks
    fg_net_optimizer.zero_grad()
    bg_net_optimizer.zero_grad()
    sp_net_optimizer.zero_grad()        
    ### optimizers for auxiliary classifiers
    fc_net_optimizer.zero_grad()
    bc_net_optimizer.zero_grad()

    ### Regularizations
    # Pseudo-label for additional regularization
    f_logits = fc_net(fg)
    b_logits = bc_net(bg)
    hpq_f = cross_ent(zf_logits, f_logits)
    hpq_b = cross_ent(zb_logits, b_logits)
    # orthogonal regularizations
    ortho_reg = orthogonal_reg(fg_net) + \
                    orthogonal_reg(bg_net)
    # TV-norm
    tv_norm = tv_loss(bg_shuf)

    ### log-lkhd for generators
    log_pf = - F.l1_loss(fg, im_t, reduction='none') 
             / (2. * SIGMA ** 2)
    log_pb = - F.l1_loss(bg_shuf, im_t, reduction='none') 
             / (2. * SIGMA ** 2)
            
    # posterior responsibility
    with torch.no_grad():
        ga_f = pi_f * log_pf.exp() 
             / (pi_f * log_pf.exp() + pi_b * log_pb.exp() + 1e-8)
    # objective for image generation
    e_z_log_p = ga_f.detach() * ((pi_f + 1e-8).log() + log_pf) \
       + (1. - ga_f.detach()) * ((pi_b + 1e-8).log() + log_pb)
                       
    G_loss = - e_z_log_p.mean() + tv_norm * .01 + \
             hpq_f * .1 + hpq_b * .1 + ortho_reg * 1

    G_loss.backward()
    fg_net_optimizer.step()
    bg_net_optimizer.step()
    sp_net_optimizer.step()
        
    fc_net_optimizer.step()
    bc_net_optimizer.step()

def update_E(en_pos, en_neg):
    ebm_optimizer.zero_grad()
        
    ebm_loss = en_pos.mean() - en_neg.mean()
    ebm_loss.backward()

    ebm_optimizer.step()

def learn(zp, zn, im_t):
    ### 1. Sampling latent variables
    zp = sample_langevin_posterior(zp)
    zn = sample_langevin_prior(zn)

    ### 2. Updating the parameters
    en_pos, zpf_logits, zpb_logits = ebm_net(zp)
    en_neg, znf_logits, znb_logits = ebm_net(zn)
    # update LEBMs
    update_E(en_pos, en_neg)

    im_p, fg, bg_shuf, pi_f, pi_b, bg = forward(zp)
    # update the generators
    update_G(im_t, fg, bg_shuf, pi_f, pi_b, bg,
             zf_logits, zb_logits)

\end{lstlisting}

\newpage
\section{Evaluation Protocols}
\paragraph{Intersecion of Union (IoU)}
The IoU score measures the overlap of two regions $A$ and $B$ by calculating the ratio of intersection over union, according to
\begin{equation}
    \text{IoU}(A, B) = \frac{|A \cap B|}{|A \cup B|},
\label{equ:iou}
\end{equation}
where we use the inferred mask and ground-truth mask as $A$ and $B$ respectively for evaluation.

\paragraph{Dice (F1) score}
Similarly, the Dice (F1) score is
\begin{equation}
    \text{Dice}(A, B) = \frac{2 |A \cap B|}{|A| + |B|}.
\label{equ:dice}
\end{equation}
Higher is better for both scores.

\begin{figure}[!htbp]
    \centering
    \includegraphics[width=\linewidth]{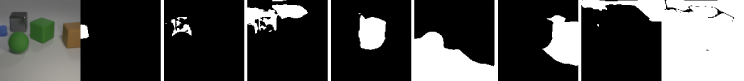}
    \caption{An example situation when using each individual mask as the background mask gives higher scores. Note that if we threshold the output of each individual slot and compose them, the result would be the mask shown in the last column.}
    \label{fig:example_iodine}
\end{figure}

\begin{figure}[!htbp]
    \centering
    \includegraphics[width=\linewidth]{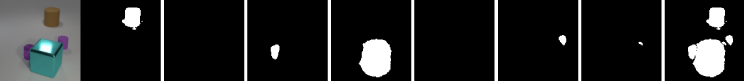}
    \caption{An example situation when thresholding\&combining the output of each individual slot gives higher scores. We can see from the last column that the combined mask fits the foreground objects well.}
    \label{fig:example_sltn}
\end{figure}

\paragraph{Evalution}
As mentioned in the paper, IODINE~\cite{greff2019multi} and Slot-attention~\cite{locatello2020object} are designed for segmenting complex multi-object scenes using slot-based object representations. Ideally, the output of these models consists of masks for each individual object, while the background is viewed as a virtual “object” as well. In practice, however, it is possible that the model distributes the background over all the slots as mentioned in~\citet{locatello2020object}. Taking both cases into consideration (see~\cref{fig:example_iodine} and~\cref{fig:example_sltn}), we propose two approaches to convert the multiple output masks into a foreground-background partition, and report the best results of these two options: (1) we compute the scores by making each mask as the background mask at a time, and then choose the best one; this works better when the background is treated as a virtual "object"; (2) we threshold and combine all the masks into a foreground mask; this is for when background is distributed to all slots.

\newpage
\section{Additional Illustrations and Baseline Results}
\subsection{More Examples} 
\begin{figure}[!htbp]
    \centering
    \includegraphics[width=\linewidth]{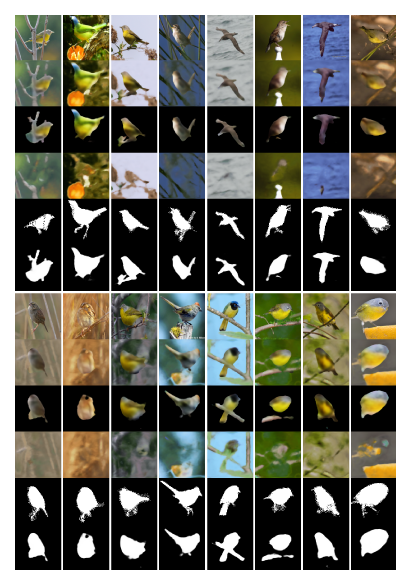}
    \caption{Additional foreground extraction results on \ac{birds} dataset.}
    \label{fig:supp_birds}
\end{figure}

\begin{figure}[!htbp]
    \centering
    \includegraphics[width=\linewidth]{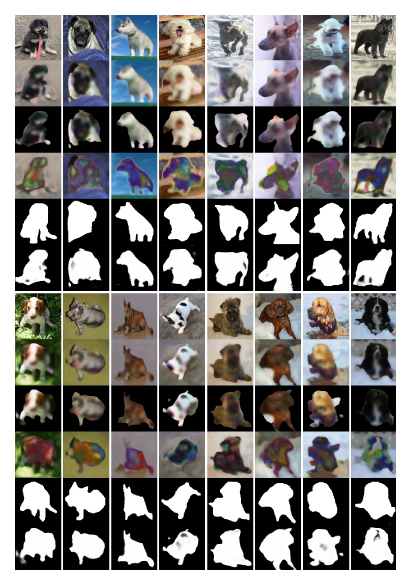}
    \caption{Additional foreground extraction results on \ac{dogs} dataset.}
    \label{fig:supp_dog}
\end{figure}

\begin{figure}[!htbp]
    \centering
    \includegraphics[width=\linewidth]{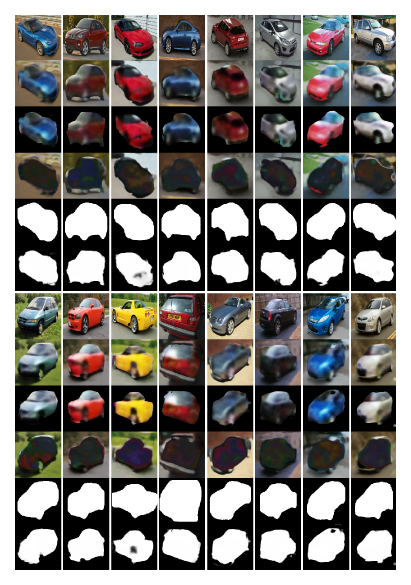}
    \caption{Additional foreground extraction results on \ac{cars} dataset.}
    \label{fig:supp_car}
\end{figure}

\begin{figure}[!htbp]
    \centering
    \includegraphics[width=\linewidth]{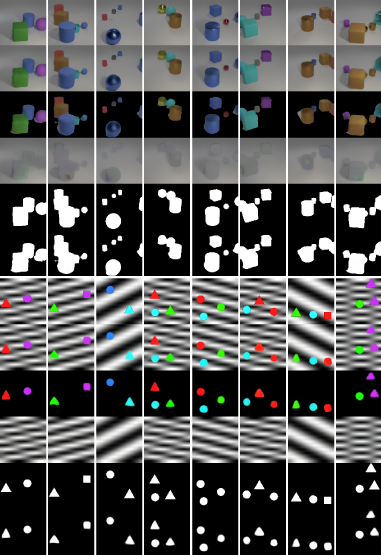}
    \caption{Additional foreground extraction results on CLEVR6 and \ac{tmds} datasets.}
    \label{fig:supp_clevr_textured}
\end{figure}

We provide more foreground extraction results of our model for each dataset; see \cref{fig:supp_birds}, \cref{fig:supp_dog}, \cref{fig:supp_car} and \cref{fig:supp_clevr_textured}. From top to bottom, we display: (i) observed images, (ii) generated images, (iii) masked generated foregrounds, (iv) generated backgrounds, (v) ground-truth foreground masks, and (vi) inferred foreground masks in each figure.

\subsection{Failure Modes}
\begin{figure}[!htbp]
    \centering
    \includegraphics[width=\linewidth]{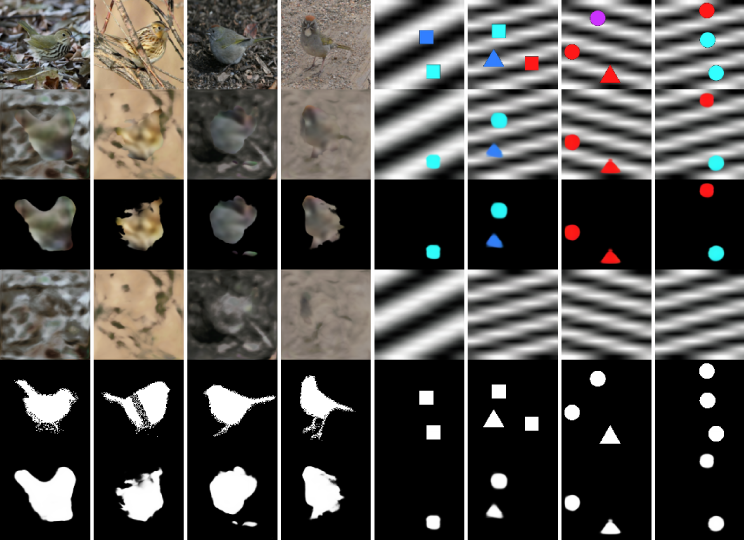}
    \caption{Typical failure modes on \ac{birds} and \ac{tmds}.}
    \label{fig:supp_failure_modes}
\end{figure}

We provide examples for illustrating typical failure modes of the proposed model; see \cref{fig:supp_failure_modes}. On \ac{birds} dataset, we observe that the method can perform worse on samples where the foreground object has colors and textures quite similar to the background regions. Although the method can still capture the rough shape of the foreground object, some details can be missing. On \ac{tmds} dataset, we observe that the method may occasionally miss one of the foreground objects. We conjecture that the problem can be mitigated with more powerful generator and further fine-tuning on this dataset.

\subsection{Baseline Results}
\begin{figure}[!htbp]
    \centering
    \includegraphics[width=\linewidth]{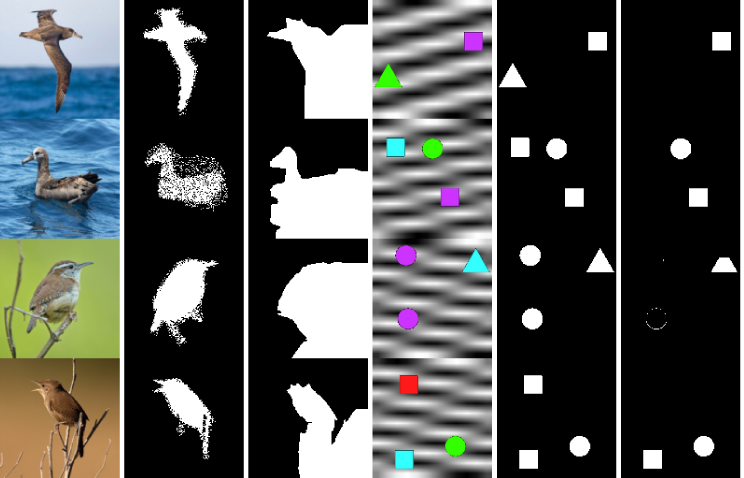}
    \caption{Results of GrabCut on \ac{birds} and \ac{tmds} datasets. The first three columns are results from \ac{birds} dataset, and the last three are from \ac{tmds}. From left to right, we display the observed image, ground-truth mask, and the foreground extraction results respectively.}
    \label{fig:supp_grabcut}
\end{figure}

\paragraph{GrabCut}
We provide results of GrabCut~\cite{rother2004grabcut} on \ac{birds} dataset and \ac{tmds} dataset, shown in \cref{fig:supp_grabcut}. We can see that GrabCut algorithm may fail when the foreground object and background region have moderately similar colors and textures. On \ac{tmds} dataset, GrabCut algorithm outperforms other baselines, but is still inferior to the proposed method and exhibits a similar failure pattern.

\begin{figure}[!htbp]
    \centering
    \includegraphics[width=\linewidth]{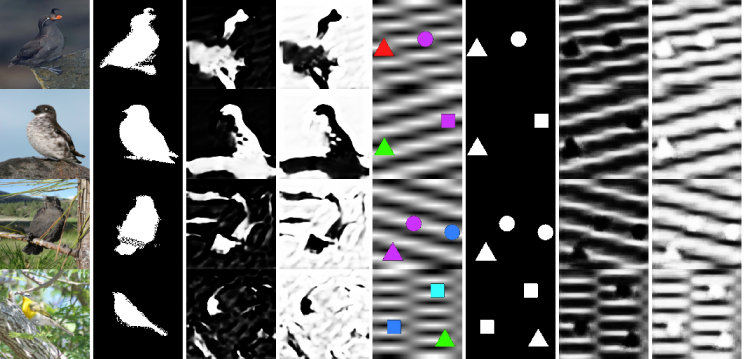}
    \caption{Results of ReDO on \ac{birds} and \ac{tmds} datasets. The first four columns are results from \ac{birds} dataset, and the last four are from \ac{tmds}. From left to right, we display the observed image, ground-truth mask, mask from the first output channel and from the second channel respectively.}
    \label{fig:supp_redo}
\end{figure}

\paragraph{ReDO}
We provide results of ReDO~\cite{chen2019unsupervised} on \ac{birds} dataset and \ac{tmds} dataset, shown in \cref{fig:supp_redo}. ReDO overall performs better than GrabCut on \ac{birds} dataset, while it may fail when the background regions become more complex. We can also observe that ReDO relies heavily on the pixel intensities for foreground-background grouping on \ac{tmds} dataset.

\begin{figure}[!htbp]
    \centering
    \includegraphics[width=\linewidth]{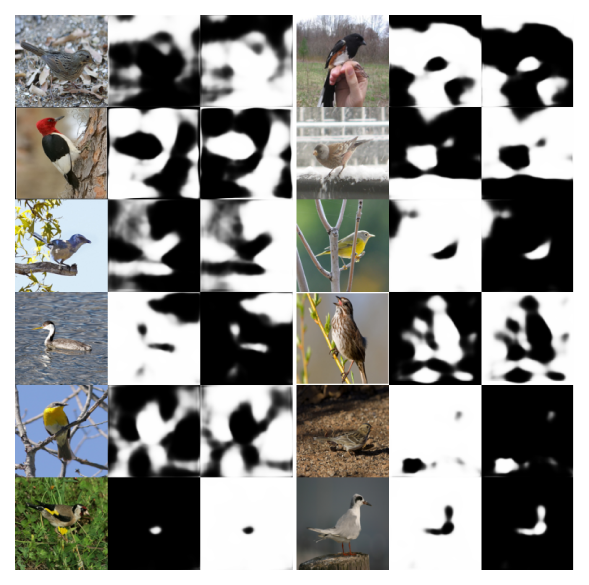}
    \caption{Results of IODINE on \ac{birds} datasets. We provide the observed image, mask from the first slot and from the second slot respectively.}
    \label{fig:supp_iodine_birds}
\end{figure}

\begin{figure}[!htbp]
    \centering
    \includegraphics[width=0.95\linewidth]{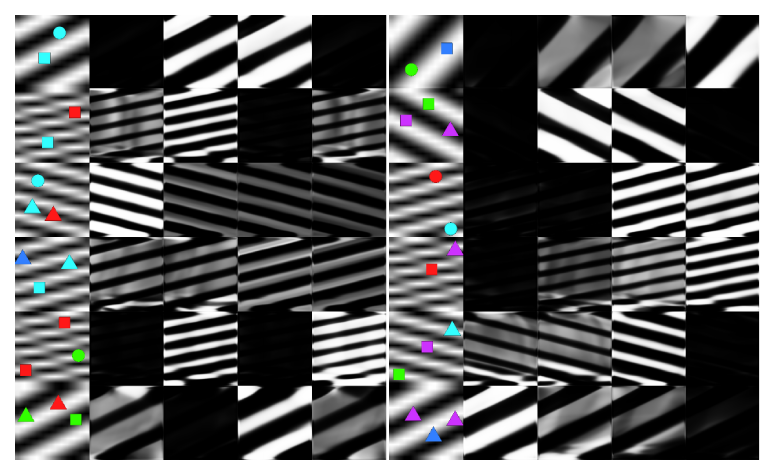}
    \caption{Results of IODINE on \ac{tmds} datasets. We provide the observed image and masks from four object slots respectively.}
    \label{fig:supp_iodine_textured}
\end{figure}

\paragraph{IODINE}
On \ac{birds} dataset, we observe that IODINE~\cite{greff2019multi} tends to use color as a strong cue for segmentation, see \cref{fig:supp_iodine_birds}. On \ac{tmds} dataset, IODINE is distracted by the background; see \cref{fig:supp_iodine_textured}. These two findings are consistent with those reported in~\cite{greff2019multi};

\begin{figure}[!htbp]
    \centering
    \includegraphics[width=0.85\linewidth]{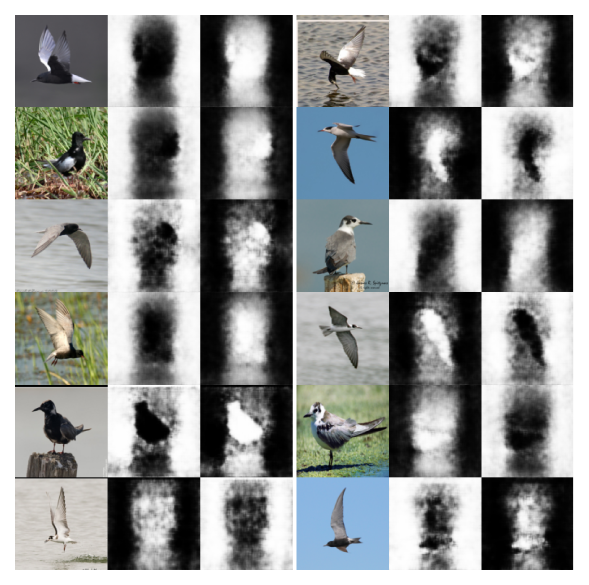}
    \caption{Results of Slot-Attention on \ac{birds} datasets. We provide the observed image, mask from the first slot and from the second slot respectively.}
    \label{fig:supp_sltn_birds}
\end{figure}

\begin{figure}[!htbp]
    \centering
    \includegraphics[width=\linewidth]{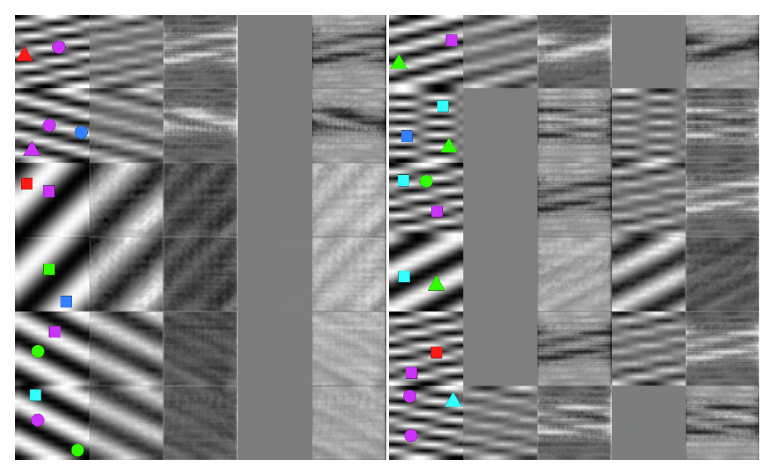}
    \caption{Results of Slot-Attention on \ac{tmds} datasets. We provide the observed image and masks from four object slots respectively.}
    \label{fig:supp_sltn_textured}
\end{figure}

\paragraph{Slot-Attention}
On \ac{birds} dataset, Slot-attention learns to roughly locate the position of foreground objects, but mostly fails to provide foreground masks when the background region becomes complex; see \cref{fig:supp_sltn_birds}. Similarly, we can observe that Slot-Attention tends to use color as a strong cue for segmentation. On \ac{tmds} dataset, Slot-attention is distracted by the background; see \cref{fig:supp_sltn_textured}.

\clearpage

\newpage
\section{Further Discussion}
\subsection{Preliminary Analysis of Real-world Datasets}
\begin{figure}[!htbp]
    \centering
    \includegraphics[width=\linewidth]{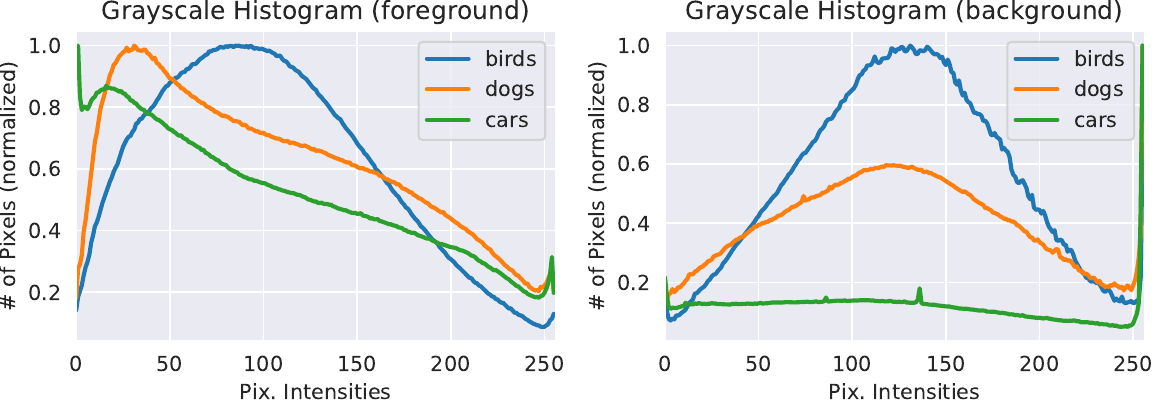}
    \caption{Image histograms for foreground objects and background regions from each dataset.}
    \label{fig:hist_supp}
\end{figure}

\begin{figure}[!htbp]
    \centering
    \includegraphics[width=\linewidth]{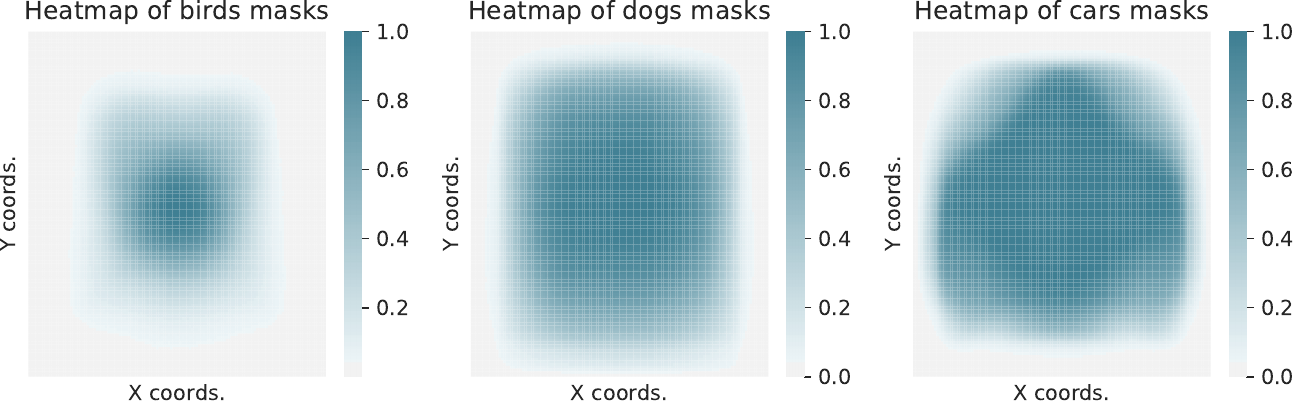}
    \caption{Heatmaps of ground-truth masks for each dataset.}
    \label{fig:heatmap_supp}
\end{figure}

We provide preliminary analysis of the statistics of the three real-world datasets. To measure the similarity of colors and textures for these datasets, we calculate the image histogram for the foreground objects and background regions of each dataset; see \cref{fig:hist_supp}. To probe the similarity of shape distributions, we also provide the heatmap of foreground masks, as shown in \cref{fig:heatmap_supp}. The heatmaps are calculated by overlapping the ground-truth masks and normalizing the summarized intensities with the maximum values. Despite the apparent difference in \ac{birds} vs \ac{dogs} and \ac{cars}, we can see that the data distribution of \ac{birds} dataset is more similar to that of \ac{dogs} dataset than to that of \ac{cars} dataset. We can also observe the similarity between the distributions of \ac{dogs} and \ac{cars} datasets. This could partly explain why the proposed method shows relatively strong performance on objects from unseen categories, \ie, it effectively combines the colors, textures and shapes for foreground extraction.

\subsection{Possible Extension to Multi-Object Segmentation}
\begin{figure}[!htbp]
    \centering
    \includegraphics[width=\linewidth]{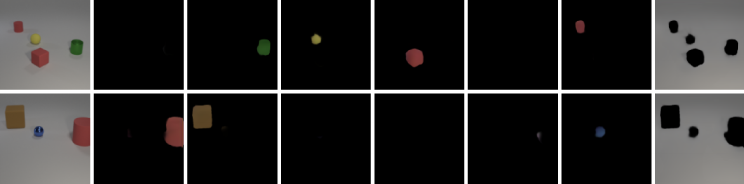}
    \caption{Preliminary results on learning slot-based object representation.}
    \label{fig:en_slt_supp}
\end{figure}

\begin{figure}[!htbp]
    \centering
    \includegraphics[width=\linewidth]{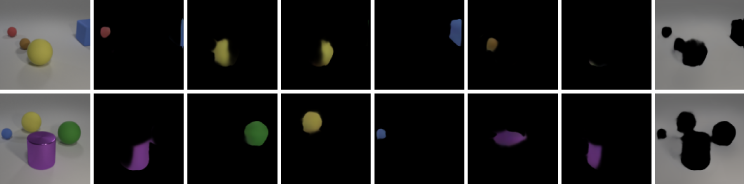}
    \caption{Failure modes of energy-based slot representation model.}
    \label{fig:en_slt_failure_supp}
\end{figure}

We explore the possibility of using our model for segmenting and disentangling multiple objects. As shown in \cref{fig:en_slt_supp}, the proposed method can disentangle the foreground objects, while providing explicit identification of the background region. However, we find that the model occasionally distributes a single object into several slots based on the difference in texture and shading; see \cref{fig:en_slt_failure_supp}. We conjecture that this is due to the lack of objectness modeling. We would like to investigate more on this direction in future works.

\subsection{Prior Sampling Results on Birds Dataset}
\begin{figure}[!htbp]
    \centering
    \includegraphics[width=\linewidth]{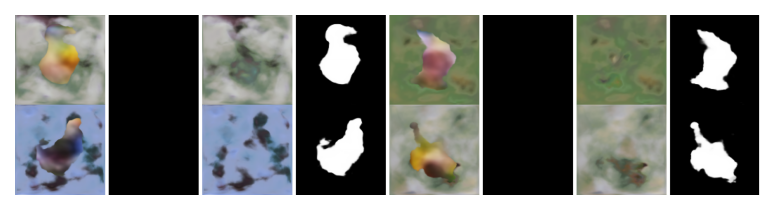}
    \caption{Prior sampling results on \ac{birds} dataset.}
    \label{fig:prior_sample_supp}
\end{figure}

We provide preliminary results of sampling from the learned energy-based priors, as shown in \cref{fig:prior_sample_supp}. Of note, the generated prior samples are generally less realistic compared with the posterior samples, as prior sampling does not involve the region competition between foreground and background components, which may lead to worse separation and the generation of foreground and background regions. We would further explore generating foreground and background in future work.

\clearpage

\end{document}